\documentclass[journal,twoside,web]{ieeecolor}
\usepackage{tmi}
\usepackage{amsmath,amssymb,amsfonts}
\usepackage{algorithmic}
\usepackage{graphicx}
\usepackage{textcomp}
\usepackage{hyperref}
\hypersetup{
  colorlinks=true,
  linkcolor=red,
  citecolor=blue 
}
\usepackage{multirow}

\usepackage{tabularx}
\usepackage{colortbl}
\usepackage{epstopdf}
\usepackage{booktabs}
\usepackage{soul}
\usepackage{booktabs,makecell}
\usepackage{diagbox}
\usepackage{arydshln}
\usepackage{paralist}
\usepackage{tablefootnote}
\usepackage{algorithm}
\usepackage{listings}

\usepackage{etoolbox}

\newif\ifinrevblock
\inrevblockfalse

\AtBeginEnvironment{figure}{\ifinrevblock\color{blue}\fi}
\AtBeginEnvironment{figure*}{\ifinrevblock\color{blue}\fi}
\AtBeginEnvironment{table}{\ifinrevblock\color{blue}\fi}
\AtBeginEnvironment{table*}{\ifinrevblock\color{blue}\fi}

\newtheorem{theorem}{Theorem}

\definecolor{commentcolor}{rgb}{0, 0.5, 0}
\definecolor{keywordcolor}{rgb}{0.5, 0, 0.5}

\definecolor{columbiablue}{rgb}{0.61, 0.87, 1.0}
\definecolor{grannysmithapple}{rgb}{0.66, 0.89, 0.63}
\definecolor{lavenderrose}{rgb}{0.98, 0.63, 0.89}

\begin{document}
\title{Mitigating Spurious Invariance in Contrastive Learning: A Probabilistic Self-Supervised Learning Framework for Medical Image Analysis}
\author{Wei Wei, and Yading Yuan
\thanks{Wei Wei and Yading Yuan are with Columbia University Irving Medical Center, New York, NY 10032, USA, and the Data Science Institute, Columbia University, New York, NY 10027, USA (e-mail: ww2715@columbia.edu; yading.yuan@columbia.edu).}
\thanks{%
\textcopyright~2026 IEEE. Personal use of this material is permitted.
Permission from IEEE must be obtained for all other uses,
in any current or future media, including reprinting/republishing
this material for advertising or promotional purposes,
creating new collective works, for resale or redistribution
to servers or lists, or reuse of any copyrighted component
of this work in other works.
This article is published in
IEEE Transactions on Medical Imaging,
vol. 45, no. 7, pp. 3862--3877, 2026.
DOI: \url{https://doi.org/10.1109/TMI.2026.3687158}.
}
}

\maketitle
\begin{abstract}
Owing to the prohibitive cost of manual annotation for enormous medical images, self-supervised learning (SSL) has gained substantial attention and shown promise in various medical imaging tasks. Among SSL approaches, contrastive learning has emerged as a prominent one, encouraging models to encode semantic information that remains invariant between different augmented views. However, this invariance can be spurious for medical images, as subtle but critical medical features can be suppressed during augmentations. In this paper, we mitigate this issue by encoding the learned feature embeddings as probability distributions rather than fixed vectors, thus capturing a richer and more nuanced spectrum of information. Furthermore, we propose a unified probabilistic framework that seamlessly integrates multiple SSL approaches. Our method achieves superior performance over the state-of-the-art SSL methods on various downstream tasks across multiple datasets, including medical image segmentation and classification. We further validate the method through additional analyses of robustness to stronger augmentations, label efficiency under limited fine-tuning labels, and computational trade-offs. The code is released at \url{https://github.com/wwzjer/ProbSSL}.

\end{abstract}


\section{Introduction}
\label{sec: introduction}

\begin{figure}
    \centering
    \includegraphics[width=0.9\linewidth]{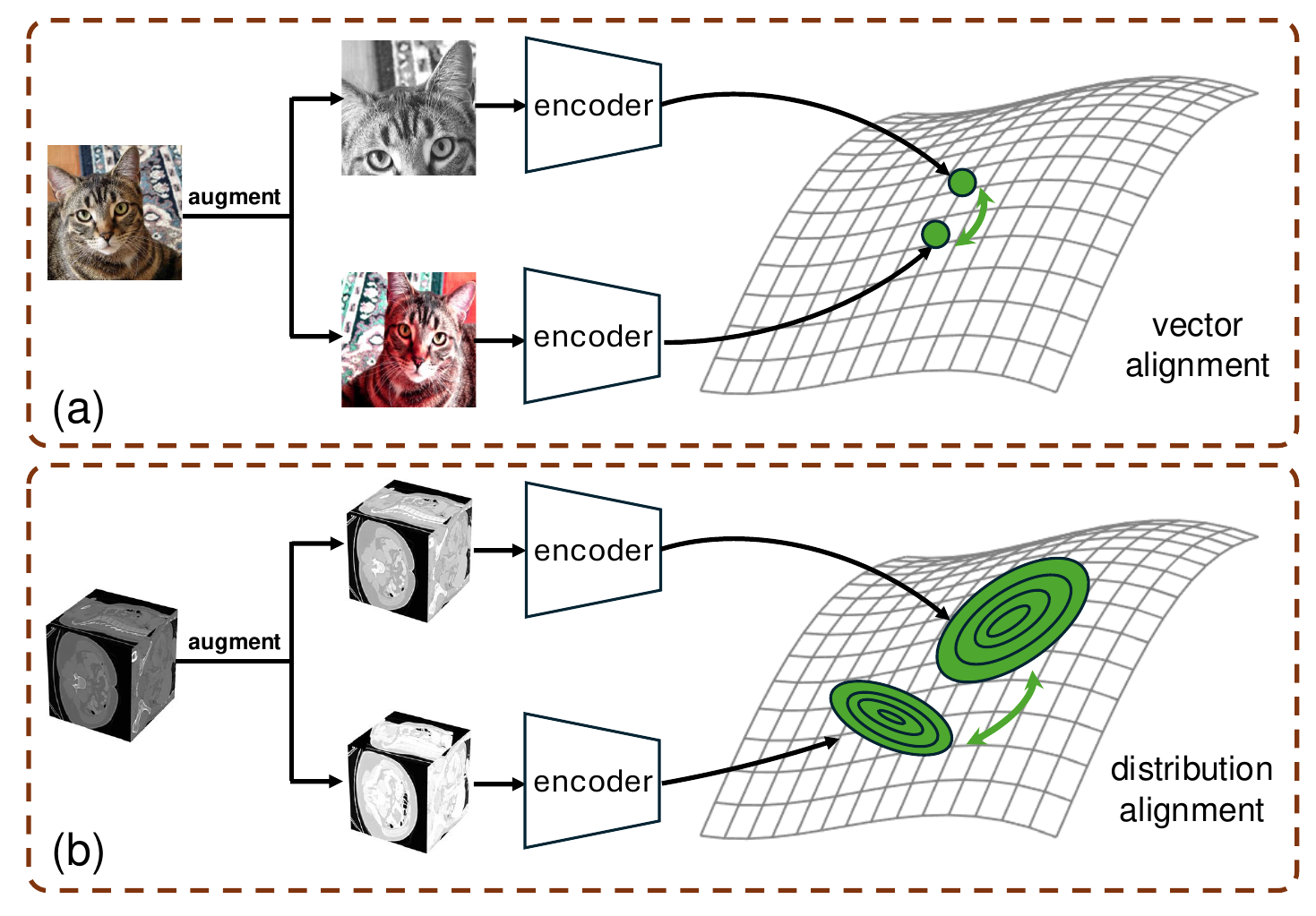}
    \caption{Motivation of our method: In (a), augmenting an input cat image typically preserves its semantic label 'cat', making it reasonable to enforce representation invariance across augmented views. In (b), however, two augmented views of a patient's CT scan may obscure crucial clinical details, causing the model to overlook shared pathology. Consequently, enforcing invariance as in scenario (a) could impair the model's capacity to learn robust, disease-relevant representations. To overcome this issue, as illustrated in (b), we represent augmented views as probability distributions instead of fixed vectors, thereby capturing richer semantic details. Accordingly, we replace conventional vector alignment in contrastive learning with distribution alignment, enhancing representation learning for medical image analysis tasks.
    }
    \label{fig1}
\end{figure}

\IEEEPARstart{R}{ecent} advances in medical imaging can be attributed to the development of foundation models~\cite{bommasani2021opportunities,moor2023foundation}, which are large-scale neural networks pre-trained on massive, diverse datasets. By leveraging the rich representations learned during pre-training, these models can be efficiently adapted to a wide range of downstream tasks with fewer annotated samples. For foundation models, self-supervised learning (SSL) is typically used, which aims to learn versatile, generalizable representations solely from unlabeled data through solving pretext tasks that are carefully designed.

Based on their pretext tasks, SSL methods can be broadly categorized into three primary approaches: i) \emph{predictive learning} aims at capturing structural information in the learned representations by predicting transformations applied to inputs, e.g., predicting rotations applied to an image~\cite{gidaris2018unsupervised}; ii) \emph{generative learning} emphasizes preserving fine-grained details by reconstructing missing parts of inputs, e.g., MAE~\cite{he2022masked}; iii) \emph{contrastive learning} focuses on learning discriminative representations by bringing positive pairs (e.g., augmentations of the same instance) together while pushing negative pairs (e.g., different instances) apart in the representation space, e.g., SimCLR~\cite{chen2020simple}.

Among SSL approaches, contrastive learning has gained significant attention in computer vision due to its strong downstream performance~\cite{chen2020simple,he2020momentum,grill2020bootstrap,chen2021exploring}. However, its application to medical imaging is not straightforward. Contrastive learning relies on the assumption that high-level semantic information remains invariant across augmented views of the same image (e.g., via random cropping, grayscale conversion, and color jitter), encouraging the model to extract these semantics into a shared representation while discarding augmentation-specific superficial nuisances~\cite{tian2020makes}. However, this assumption may be problematic in medical image analysis. For example, spatial transformations can distort critical anatomical structures, and intensity perturbations can obscure subtle pathological signals~\cite{huang2023self}. In such cases, rigidly enforcing invariance may lead to \emph{spurious invariance}, where the model fails to capture the essential medical cues in the shared representations, inadvertently suppressing task-relevant features, ultimately leading to suboptimal performance in downstream tasks like medical segmentation and diagnosis.

To address this problem, one intuitive idea is to develop high-quality augmentations tailored to medical images, guided by medical expertise~\cite{wang2024crafting}. However, we argue that it has two significant limitations. First, it relies heavily on specialized medical domain knowledge, which may not always be readily accessible. Second, self-supervised pre-training is inherently agnostic to downstream tasks. Even with meticulous augmentation engineering, the learned representation is not guaranteed to be universally beneficial for all downstream tasks. 

In this paper, we take a different route to mitigate spurious invariance by enriching the learned representations through introducing a probabilistic SSL framework. Unlike conventional SSL methods that yield deterministic embeddings, our method generalizes these methods by representing each medical image as a probabilistic distribution, with the mean capturing shared high-level features and the variance reflecting augmentation-induced variability, offering greater flexibility. Specifically, we introduce a distribution-wise contrastive loss that aligns both means and variances for a positive pair, enabling a dynamic alignment strategy where pairs exhibiting higher variability receive attenuated alignment enforcement. By relaxing rigid alignment, our method mitigates the risk of discarding critical task-relevant anatomical or pathological features, enhancing the utility of learned representations, and improving performance across diverse downstream applications.

Building on the goal of mitigating spurious invariance problem for contrastive SSL, the proposed probabilistic framework seamlessly integrates with the other two SSL approaches: predictive learning and generative learning. We employ the reparameterization trick~\cite{kingma2014autoencoding} to sample a vector from the learned probabilistic distribution and then pass it through both a classification head for predicting transformations (e.g., rotations) and a reconstruction head to re-generate the original input. This integrated design ensures that our probabilistic SSL framework supports both predictive and generative objectives, complementing the contrastive learning process. By unifying these objectives within a probabilistic framework, the model learns more expressive and robust representations, making it highly effective for downstream tasks in medical imaging.

To demonstrate the effectiveness of the proposed probabilistic SSL framework, we pre-train our method on a large, unlabeled dataset of CT scans. After pre-training, we evaluate the utility of learned representations by transferring them to downstream supervised tasks, including CT/MR image segmentation and classification. Experimental results show that our framework outperforms the baselines and achieves superior performance compared to state-of-the-art self-supervised learning methods. Beyond aggregate accuracy gains, we further validate our framework by showing improved robustness under more aggressive augmentations, better label efficiency in low-annotation regimes, and favorable computational efficiency against strong medical SSL baselines.

In summary, our contributions are five-fold:

\begin{compactitem}
    \item We pioneerly investigate the spurious invariance problem when applying contrastive learning in medical image analysis.
    \item We propose a probabilistic contrastive learning method to handle this problem, with a dynamic alignment scheme.
    \item We provide theoretical analysis to prove that the proposed probabilistic contrastive learning loss incurs a smaller estimation error, explicitly mitigating spurious invariance.
    \item We present a unified probabilistic SSL framework that integrates multiple SSL approaches to effectively learn representations for volumetric medical images.
    \item We validate the effectiveness of the proposed probabilistic framework through comprehensive downstream tasks, including CT/MR image segmentation and classification.
\end{compactitem}

\section{Related Work}
\label{sec: related_work}

Since this paper focuses on image-based inputs, we first summarize SSL approaches applied to natural image inputs in computer vision (Sec.~\ref{sec: 2A}), followed by their adaptation for 3D medical imaging (Sec.~\ref{sec: 2B}), and finally, probabilistic methods related to ours (Sec.~\ref{sec: 2C}).

\subsection{Self-supervised Learning in Computer Vision}
\label{sec: 2A}

SSL has emerged as a powerful paradigm for leveraging large-scale unlabeled data in computer vision. Representative SSL methods can be categorized into \emph{predictive}, \emph{generative}, and \emph{contrastive} learning approaches. \emph{Predictive learning} involves solving tasks where the model predicts specific transformations applied to the input, such as predicting rotation~\cite{gidaris2018unsupervised}, colorization~\cite{larsson2017colorization}, and solving jigsaw puzzles~\cite{noroozi2016unsupervised}. \emph{Generative learning}, on the other hand, focuses on recovering missing parts of the input, exemplified by tasks like inpainting~\cite{pathak2016context} or masked image modeling~\cite{he2022masked}.

As a key focus in SSL, \emph{contrastive learning} seeks to create multiple views of the same input through augmentations and learn invariant representations across these views, which captures high-level semantics. Prominent methods such as SimCLR~\cite{chen2020simple} and MoCo~\cite{he2020momentum} achieve this by distinguishing positive pairs (two augmented views of the same image) from negative pairs (different images). Recently, non-contrastive methods have eliminated the need for negative pairs by only maximizing agreement between the representations of positive pairs. To avoid representation collapse, they adopt different regularization strategies, either through their training objectives~\cite{zbontar2021barlow,bardes2022vicreg} or specific network architectures~\cite{grill2020bootstrap,chen2021exploring,caron2021emerging}.

\subsection{Self-supervised Learning for 3D Medical Imaging}
\label{sec: 2B}

Inspired by the success of SSL in computer vision, there is growing interest in adapting these approaches to 3D medical image analysis, despite the unique challenges posed by this domain. \emph{Predictive learning} methods such as ~\cite{zhuang2019self,zhu2020rubik} design tasks like solving a Rubik's cube to predict spatial correlations between different local regions within the images. For \emph{generative learning}, methods such as ~\cite{chen2019self,zhou2021models,zhou2021preservational} utilize encoder-decoder architectures to reconstruct altered volumes back to their original state, while~\cite{chen2023masked} employs masked image modeling to recover masked volumetric information. In \emph{contrastive learning},~\cite{chaitanya2020contrastive} explores both global contrast across entire volumetric medical images and local contrast among regions within these volumes, and~\cite{zheng2022msvrl} extend non-contrastive learning to multi-level representations by incorporating a cross-level consistency loss. Furthermore, hybrid methods integrate various SSL approaches into a unified multi-task learning framework to enhance representation learning~\cite{taleb20203d,haghighi2021transferable,haghighi2022dira,tang2022self,jiang2023anatomical,wang2023swinmm,he2023geometric,wu2024voco}.

However, limited attention has been paid to the problem of spurious invariance when applying contrastive learning to medical imaging, and few studies have explored probabilistic modeling within the SSL framework for medical imaging. To the best of our knowledge, our work is among the first to recognize this significant problem and accordingly propose a probabilistic framework to handle it.

\subsection{Probabilistic Modeling for Self-supervised Learning}
\label{sec: 2C}

The integration of probabilistic modeling into SSL has emerged as a pivotal direction to advance representation learning. For instance, \cite{nakamura2023representation} formalizes SSL through variational inference to learn uncertainty-aware representations, while \cite{kirchhof2023probabilistic} enhances contrastive learning by recovering aleatoric uncertainty in ambiguous inputs. \cite{duan2024contrastive} also uses Gaussian embeddings to perform a standard probabilistic contrastive baseline, whereas our method introduces a distribution-wise contrastive loss to realize dynamic alignment upon it. Building on information theory, \cite{janiak2024enhancing} leverages probabilistic embeddings to improve out-of-distribution detection, and \cite{bizeul2024a} unifies SSL frameworks via a generative latent variable model. Meanwhile, \cite{vahidi2024probabilistic} addresses representation collapse by minimizing proper scoring rules to enhance probabilistic representation quality. Beyond theoretical advances, probabilistic SSL has demonstrated success in applications including video understanding~\cite{park2022probabilistic}, long-tailed recognition~\cite{du2024probabilistic}, and material property prediction~\cite{wang2024self}.

The novelty of our work lies in three aspects: i) We tackle the spurious invariance problem induced by aggressive augmentations in contrastive learning, a challenge usually overlooked by existing probabilistic SSL methods. ii) We propose a dynamic alignment scheme that relaxes rigid invariance constraints, enabling more flexible and robust representation learning. iii) To the best of our knowledge, this is the first probabilistic SSL framework tailored for medical volumetric representation learning.

\section{Method}
\label{sec: method}

\begin{figure*}
    \centering
    \includegraphics[width=\linewidth]{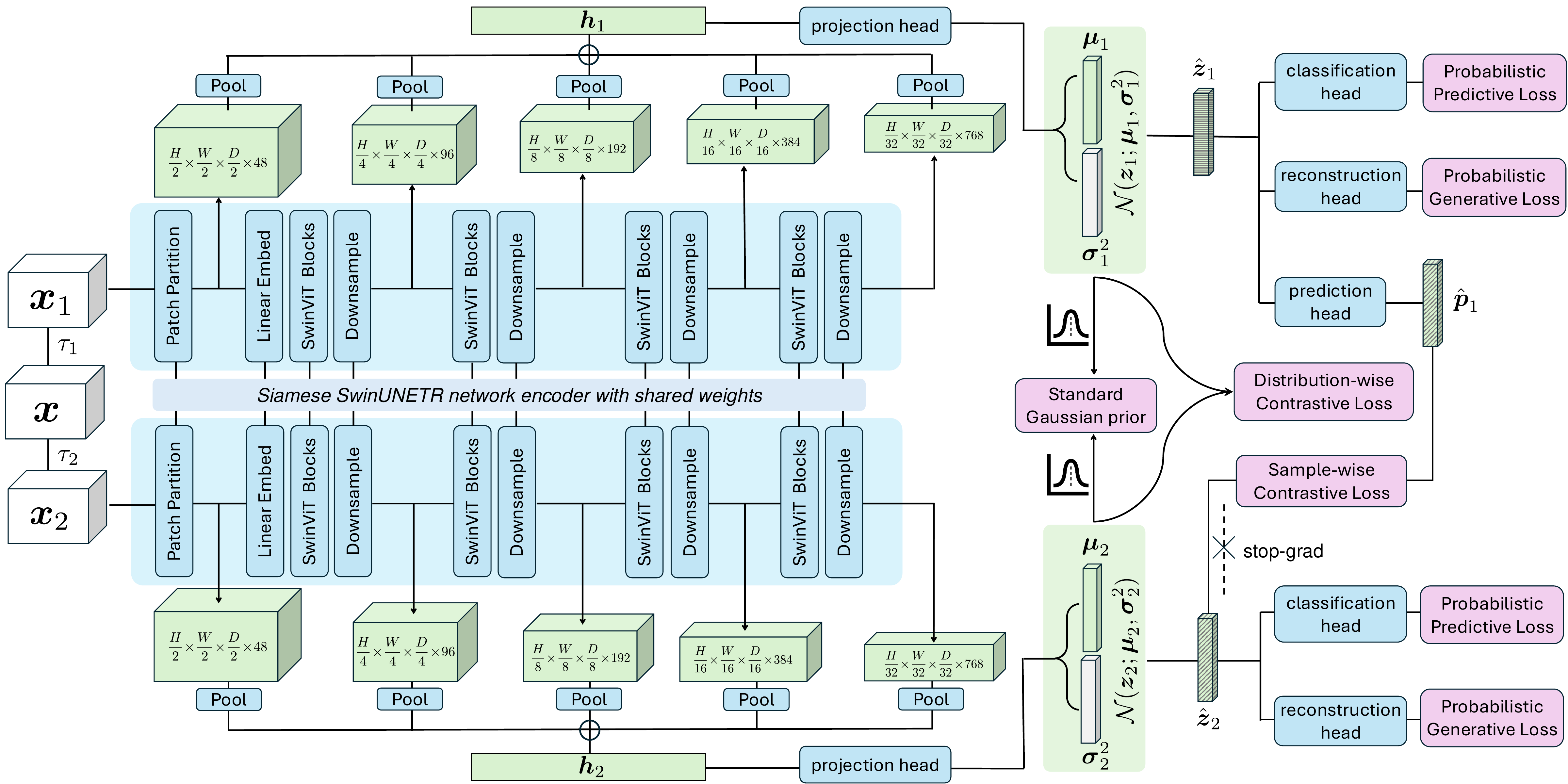}
    \caption{The entire pipeline of our probabilistic SSL framework. To improve clarity, we use distinct colors to represent different components: \textcolor{columbiablue}{network architectures}, \textcolor{grannysmithapple}{network features}, and \textcolor{lavenderrose}{training objectives}.}
    \label{fig2}
\end{figure*}

Given a large set of unlabeled volumetric medical images $\mathcal{D} = \{\boldsymbol{x}^{(i)}\}_{i=1}^N$, where $\boldsymbol{x}^{(i)}\in\mathbb{R}^{H \times W \times D}$, $H$, $W$, and $D$ denote the height, width, and depth, respectively, our goal is to train a neural network to map the input volumes to their representations that can be effectively transferred to different downstream supervised learning tasks, such as medical image segmentation and classification.

In this section, we first present the preliminaries of contrastive learning in Sec.~\ref{sec: 3A}, which motivate us to propose probabilistic contrastive learning detailed in Sec.~\ref{sec: 3B}. We then provide a theoretical analysis of our probabilistic contrastive learning method in Sec.~\ref{sec: 3C}. Finally, we extend our method by integrating it with two additional SSL approaches, leading to a unified probabilistic SSL framework described in Sec.~\ref{sec: 3D}. The complete framework is illustrated in Fig.~\ref{fig2}.

\subsection{Preliminaries of Contrastive Learning}
\label{sec: 3A}

As a prominent SSL approach, contrastive learning enables models to learn discriminative representations by contrasting positive pairs (e.g., augmented views of the same image) against negative pairs (e.g., augmented views of different images). By aligning representations of the same image under different augmentations, it extracts shared content across these augmented views, recognizing it as semantic information that remains invariant despite augmentation differences.

Specifically, given an input $\boldsymbol{x}$, two transformation functions, $\tau_1$ and $\tau_2$, are randomly drawn from a transformation distribution $\mathcal{T}$, generating two augmented views with overlap, denoted as $\boldsymbol{x}_1=\tau_1(\boldsymbol{x})$ and $\boldsymbol{x}_2=\tau_2(\boldsymbol{x})$. These augmented views are considered as a positive pair and subsequently processed by the Siamese network encoder with shared weights, where each branch comprises a backbone network $f$ to extract feature representations $\boldsymbol{h}_1=f(\boldsymbol{x}_1)$ and $\boldsymbol{h}_2=f(\boldsymbol{x}_2)$, and a projection head $g$ that maps the representations to normalized embeddings $\boldsymbol{z}_1 = g(\boldsymbol{h}_1)$ and $\boldsymbol{z}_2 = g(\boldsymbol{h}_2)$. 


A well-known example is SimCLR~\cite{chen2020simple}, which encourages the positive pairs to exhibit high similarity in the embedding space while pushing the negative pairs apart. However, it relies on large batch sizes to maintain a sufficient number of negative samples, making it particularly challenging for memory-intensive volumetric medical images. Moreover, the inherent anatomical similarity in medical scans complicates the identification of truly distinct negatives~\cite{jiang2023anatomical}. 

In recent years, a variant of contrastive learning has eliminated the dependency on negative samples~\cite{grill2020bootstrap,chen2021exploring}, making it more convenient for volumetric medical datasets. To avoid the trivial solution where all inputs collapse to constant representations, these methods employ asymmetric architectural designs. For example, SimSiam~\cite{chen2021exploring} introduces an additional prediction head $h$ that transforms one of the embeddings from the positive pair (say $\boldsymbol{z}_1$) to a prediction $\boldsymbol{p_1}$ of the other embedding $\boldsymbol{z}_2$ by $\boldsymbol{p}_1=h(\boldsymbol{z}_1)$ and $\boldsymbol{p}_2=h(\boldsymbol{z}_2)$. Additionally, a stop-gradient (\texttt{sg}) operation is applied in the learning objective to further prevent collapse by avoiding gradient flow through the target embeddings. The objective is as follows:
\begin{equation}
    \ell_{\text{SimSiam}} = -\frac{1}{2} \left(\text{sim}(\boldsymbol{p}_1, \texttt{sg}(\boldsymbol{z}_2)) + \text{sim}(\boldsymbol{p}_2, \texttt{sg}(\boldsymbol{z}_1))  \right),
\end{equation}
where the sim function between two embeddings is measured using cosine similarity: $\text{sim}(\boldsymbol{u}, \boldsymbol{v}) = \boldsymbol{u}^\top \boldsymbol{v} / (\|\boldsymbol{u}\| \|\boldsymbol{v}\|)$. By minimizing the negative cosine similarity between the predicted embedding of one view and the embedding of the other view, SimSiam encourages the backbone to output representations that are invariant between augmented views. 

The applied augmentations typically include random cropping, color jitter, Gaussian blurring, and so on for natural images, as they usually do not alter the semantic content of the image. However, as discussed in Sec.~\ref{sec: introduction} and illustrated in Fig.~\ref{fig1}, this is less suitable for medical imaging, where diagnoses often rely on subtle but crucial features that augmentations could obscure or modify, thus enforcing invariant representations might induce suboptimal results. Next, we will introduce our probabilistic contrastive learning method to mitigate this issue. 

\subsection{Probabilistic Contrastive Learning}
\label{sec: 3B}

The proposed probabilistic contrastive learning method generalizes the aforementioned conventional contrastive learning methods by modeling the embeddings probabilistically. This enables richer representational capacity by not only capturing semantic similarity across augmented views through the means, but also accounting for the variability introduced by specific augmentations through the variances.

To handle volumetric medical images, we utilize the encoder of the widely used SwinUNETR architecture~\cite{hatamizadeh2021swin} as the backbone $f$ in our framework, as illustrated in Fig.~\ref{fig2}. It processes the input volume $\boldsymbol{x}$ and extracts five hierarchical features with different resolutions, corresponding to different stages of Swin Transformer~\cite{liu2021swin}. These multi-scale features are integrated through pooling and concatenation operations to form a unified representation $\boldsymbol{h}$. The representation is then passed through a projection head $g$ that generates the embedding $\boldsymbol{z}$. Instead of treating $\boldsymbol{z}$ as a deterministic vector, we model it as a latent variable with distribution $p(\boldsymbol{z}|\boldsymbol{x})$. 

To infer the latent distribution $p(\boldsymbol{z}|\boldsymbol{x})$, we adopt the variation inference approach. Specifically, the projection head $g$ is designed with two branches that produce the mean
$\boldsymbol{\mu}$ and the variance $\boldsymbol{\sigma}^2$, forming the variational distribution $q(\boldsymbol{z}|\boldsymbol{x})$, modeled as a Gaussian $\mathcal{N}(\boldsymbol{z};\boldsymbol{\mu}, \boldsymbol{\sigma}^2)$ for its mathematical tractability. The objective is to minimize the KL-divergence between the true distribution $p(\boldsymbol{z}|\boldsymbol{x})$ and the approximated variational distribution $q(\boldsymbol{z}|\boldsymbol{x})$, which is equivalent to maximizing the Evidence Lower Bound (ELBO):
\begin{equation}
    \text{ELBO} = \mathbb{E}_{q(\boldsymbol{z}|\boldsymbol{x})} \left[ \log p(\boldsymbol{x}|\boldsymbol{z}) \right] - \text{KL}(q(\boldsymbol{z}|\boldsymbol{x}) \| p(\boldsymbol{z})).
    \label{eq: elbo}
\end{equation}
This is analogous to VAE~\cite{kingma2014autoencoding}, where the first term ensures that the learned embeddings effectively reconstruct original data, while the second term regularizes the latent distribution to align with the prior $p(\boldsymbol{z})$, which is typically a standard Gaussian distribution, preventing degenerate solutions where the learned variance collapses to zero or diverges. 

In our context, the first term does not necessarily have to reconstruct $\boldsymbol{x}$ directly. Its broader goal is to ensure that $\boldsymbol{z}$ contains meaningful information about $\boldsymbol{x}$. This goal can be satisfied by other objectives that promote the informativeness of $\boldsymbol{z}$, such as aligning representations of augmented views in contrastive learning. Here we denote $\boldsymbol{x}_1$ and $\boldsymbol{x}_2$ as two augmented views of the same medical image $\boldsymbol{x}$. By aligning their probabilistic embeddings $\boldsymbol{z}_1\sim q(\boldsymbol{z}_1|\boldsymbol{x}_1)=\mathcal{N}(\boldsymbol{z}_1;\boldsymbol{\mu}_1, \boldsymbol{\sigma}_1^2)$ and $\boldsymbol{z}_2\sim q(\boldsymbol{z}_2|\boldsymbol{x}_2)=\mathcal{N}(\boldsymbol{z}_2;\boldsymbol{\mu}_2, \boldsymbol{\sigma}_2^2)$, the encoder learns to extract invariant features from $\boldsymbol{x}$. This alignment serves as a proxy for reconstruction, as it ensures that $\boldsymbol{z}$ retains key information about the data.

To align the distributions of probabilistic embeddings, we minimize the symmetric Kullback-Leibler (KL) divergence between the Gaussian distributions of positive pairs, as a Distribution-wise Contrastive Learning (DCL) loss:
\begin{equation}
    \ell_\text{DCL}= \frac{\text{KL}(q(\boldsymbol{z}_1|\boldsymbol{x}_1) \| q(\boldsymbol{z}_2|\boldsymbol{x}_2)) + \text{KL}(q(\boldsymbol{z}_2|\boldsymbol{x}_2) \| q(\boldsymbol{z}_1|\boldsymbol{x}_1))}{2}, 
\end{equation}
where 
\begin{equation}
\mathrm{KL}\big(q_1 \,\|\, q_2\big)
= \frac{1}{2}\sum_{j=1}^d
\Big[
\log\frac{\sigma_{2j}^2}{\sigma_{1j}^2}
+\frac{\sigma_{1j}^2 + (\mu_{1j}-\mu_{2j})^2}{\sigma_{2j}^2}
-1
\Big].
\end{equation}
Here, $\mu_{\cdot j}$ denotes the $j$-th element of the mean vectors, while $\sigma_{\cdot j}$ denotes the standard deviations. This assumes diagonal covariances $\boldsymbol{\Sigma}_{\cdot}=\mathrm{diag}(\sigma_{\cdot 1}^{2},\ldots,\sigma_{\cdot d}^{2})$.

Combining them together, the log terms are canceled, and DCL loss admits the following closed form:
\begin{equation}
\ell_{\text{DCL}}
= \frac{1}{4}\sum_{j=1}^d
\Big[
\big(\mu_{1j}-\mu_{2j}\big)^2
\Big(\frac{1}{\sigma_{1j}^2}+\frac{1}{\sigma_{2j}^2}\Big)
+
\Big(\frac{\sigma_{1j}^2}{\sigma_{2j}^2}
+\frac{\sigma_{2j}^2}{\sigma_{1j}^2}
-2\Big)
\Big].
\label{eq:DCL}
\end{equation}
In this loss, the first term encourages the means $\boldsymbol{\mu}_1$ and $\boldsymbol{\mu}_2$ to align,  weighted by the sum of their inverse variances, while the second term achieves its minimum value of zero when the variances match. In conventional contrastive learning, where the probability mass collapses into a deterministic point, enforcing invariance by aligning deterministic embeddings of different augmented views can overlook critical medical cues that are obscured by specific augmentations. In contrast, our formulation which is a dynamic alignment strategy acts as a buffer: if augmentations produce greater variability, the mean alignment in the first term is naturally attenuated, reducing the risk of over-aligning noisy augmented views. 

We emphasize that the derived formulation of the \textbf{DCL objective is novel within the self-supervised learning literature}, as it yields a \textbf{dynamic alignment mechanism}. Specifically, in Eq.~\ref{eq:DCL}, each latent dimension $j$ contributes a term $(\mu_{1j}-\mu_{2j})^2\!\left(\tfrac{1}{\sigma_{1j}^2}+\tfrac{1}{\sigma_{2j}^2}\right)$ which adaptively scales the alignment strength by the inverse variances. Consequently, dimensions with higher augmentation-induced variability (larger $\sigma_{ij}^2$) exert weaker alignment pressure,  while more reliable dimensions (smaller $\sigma_{ij}^2$) receive stronger enforcement. This fine-grained control across latent dimensions effectively mitigates spurious invariance in medical imaging. Therefore, by aligning probabilistic distributions rather than deterministic points, DCL is more robust to noise and variability induced in the data augmentation. 

To prevent representation collapse, inspired by SimSiam~\cite{chen2021exploring}, a prediction head $h$ is introduced to map the sample from the first embedding $q(\boldsymbol{z}_1|\boldsymbol{x}_1)$ to the prediction of the sample from the second distribution $q(\boldsymbol{z}_2|\boldsymbol{x}_2)$, and vice versa. Here, reparameterization trick~\cite{kingma2014autoencoding} can be utilized to realize this, where we sample latent vectors $\hat{\boldsymbol{z}}_1$ and $\hat{\boldsymbol{z}}_2$ from their distributions:
\begin{align}
    \hat{\boldsymbol{z}}_1 &= \boldsymbol{\mu}_1 + \boldsymbol{\sigma}_1 \odot \boldsymbol{\epsilon}_1, \quad \boldsymbol{\epsilon}_1 \sim \mathcal{N}(\boldsymbol{0}, \boldsymbol{I}), \notag\\
    \hat{\boldsymbol{z}}_2 &= \boldsymbol{\mu}_2 + \boldsymbol{\sigma}_2 \odot \boldsymbol{\epsilon}_2, \quad \boldsymbol{\epsilon}_2 \sim \mathcal{N}(\boldsymbol{0}, \boldsymbol{I}),
    \label{eq: reparam}
\end{align}
where $\odot$ denotes the element-wise product. This ensures that gradients can backpropagate through the stochastic sampling process. The sampled latent vectors $\hat{\boldsymbol{z}}_1$ and $\hat{\boldsymbol{z}}_2$ are passed through the prediction head $h$. The objective aligns the output of prediction head $\hat{\boldsymbol{p}}_1 = h(\hat{\boldsymbol{z}}_1)$ with $\hat{\boldsymbol{z}}_2$ and vice versa, as a Sample-wise Contrastive Learning (SCL) loss:
\begin{equation}
    \ell_{\text{SCL}} = -\frac{1}{2} \left(\text{sim}(\hat{\boldsymbol{p}}_1, \texttt{sg}(\hat{\boldsymbol{z}}_2)) + \text{sim}(\hat{\boldsymbol{p}}_2, \texttt{sg}(\hat{\boldsymbol{z}}_1))  \right).
\end{equation}

Furthermore, grounded by the second term in ELBO (Eq.~\ref{eq: elbo}), a prior term is introduced to avoid degenerate solutions:
\begin{align}
    \ell_{\text{prior}} &= \frac{1}{2}\left(\text{KL}(q(\boldsymbol{z}_1|\boldsymbol{x}_1) \| p(\boldsymbol{z})) + \text{KL}(q(\boldsymbol{z}_2|\boldsymbol{x}_2) \| p(\boldsymbol{z}))\right) \notag \\
    &= \frac{\boldsymbol{\mu}_1^2+\boldsymbol{\mu}_2^2}{4} + \frac{\boldsymbol{\sigma}_1^2+\boldsymbol{\sigma}_2^2}{4} -\frac{\log{\boldsymbol{\sigma}^2_1}+\log{\boldsymbol{\sigma}^2_2}}{4} -\frac{1}{2}.
\end{align}

The total loss of the proposed probabilistic contrastive learning (ProbCL) combines DCL, SCL and the prior term:
\begin{equation}
    \ell_\text{ProbCL} = \alpha_1 \ell_{\text{DCL}} + \alpha_2 \ell_{\text{SCL}} + \alpha_3 \ell_{\text{prior}},
    \label{eq: probcl}
\end{equation}
where $\alpha_1,\alpha_2$ and $\alpha_3$ are balancing parameters.

\subsection{Theoretical Analysis}
\label{sec: 3C}

Next, we provide a theoretical analysis that explains how our probabilistic contrastive objective operates and why it yields benefits that interpret the improved downstream performance.

Recalling the aforementioned symbols, let two augmented views $\boldsymbol{x}_1,\boldsymbol{x}_2$ of the same input $\boldsymbol{x}$ produce probabilistic embeddings $q(\boldsymbol{z}_i|\boldsymbol{x}_i)\sim\mathcal N(\boldsymbol{\mu}_i,\ \boldsymbol{\Sigma}_i), i\in\{1,2\}$. Here, we use $\boldsymbol{\Sigma}_i$ as a compact covariance notation for theoretical derivation convenience. We adopt a locally linear signal–noise decomposition of the embedding means:
\begin{equation}
\boldsymbol{\mu}_i=\boldsymbol{s}+\boldsymbol{\varepsilon}_i,\qquad \boldsymbol{\varepsilon}_i\sim \mathcal N(\boldsymbol{0},\ \boldsymbol{\Omega}_i),
\end{equation}
where \(\boldsymbol{s}\in\mathbb R^d\) is the augmentation-invariant latent code to be recovered. The perturbation $\boldsymbol{\varepsilon}_i$ captures the augmentation-specific deviation induced by augmentations with its variance \(\boldsymbol{\Omega}_i\) quantifies across-view deviation of \(\boldsymbol{\mu}_i\) around \(\boldsymbol{s}\).

For the deterministic baseline, maximizing cosine similarity between \(\boldsymbol{\mu}_1\) and \(\boldsymbol{\mu}_2\) (with unit-normalized embeddings) is equivalent, up to constants, to minimizing \(\frac{1}{2}\sum_{i=1}^2\|\boldsymbol{s}-\boldsymbol{\mu}_i\|_2^2\), which yields
\begin{align}
\hat{\boldsymbol{s}}_{\mathrm{det}}
&= \arg\min_{\boldsymbol{s}} \frac{1}{2}\sum_{i=1}^2\|\boldsymbol{s}-\boldsymbol{\mu}_i\|^2 \\
\hat{\boldsymbol{s}}_{\mathrm{det}} &= \frac{1}{2}(\boldsymbol{\mu}_1+\boldsymbol{\mu}_2),
\quad
\mathrm{Cov}(\hat{\boldsymbol{s}}_{\mathrm{det}})=\frac{1}{4}(\boldsymbol{\Omega}_1+\boldsymbol{\Omega}_2).
\end{align}

The distribution-wise contrastive loss (DCL) contains a mean-alignment term that we rewrite in its matrix form:
\begin{equation}
\ell_{\mathrm{DCL}}^{(\boldsymbol{\mu})}=\frac{1}{2}(\boldsymbol{\mu}_1-\boldsymbol{\mu}_2)^\top\big(\boldsymbol{\Sigma}_1^{-1}+\boldsymbol{\Sigma}_2^{-1}\big)(\boldsymbol{\mu}_1-\boldsymbol{\mu}_2),
\end{equation}
together with a variance-alignment term that encourages \(\boldsymbol{\Sigma}_1\) and \(\boldsymbol{\Sigma}_2\) to match. Introduce the auxiliary weighted least-squares (WLS) objective:
\begin{equation}
F(\boldsymbol{s})=(\boldsymbol{\mu}_1-\boldsymbol{s})^\top\boldsymbol{\Sigma}_1^{-1}(\boldsymbol{\mu}_1-\boldsymbol{s})+(\boldsymbol{\mu}_2-\boldsymbol{s})^\top\boldsymbol{\Sigma}_2^{-1}(\boldsymbol{\mu}_2-\boldsymbol{s}).
\end{equation}
Differentiating and setting the gradient to zero gives
\begin{equation}
\nabla_{\boldsymbol{s}} F(\boldsymbol{s})=-2\boldsymbol{\Sigma}_1^{-1}(\boldsymbol{\mu}_1-\boldsymbol{s})-2\boldsymbol{\Sigma}_2^{-1}(\boldsymbol{\mu}_2-\boldsymbol{s})=0,
\end{equation}
hence the precision-weighted centroid:
\begin{equation}
\boldsymbol{s}^\star_{\text{prob}}=\big(\boldsymbol{\Sigma}_1^{-1}+\boldsymbol{\Sigma}_2^{-1}\big)^{-1}\big(\boldsymbol{\Sigma}_1^{-1}\boldsymbol{\mu}_1+\boldsymbol{\Sigma}_2^{-1}\boldsymbol{\mu}_2\big),
\end{equation}
Substituting \(\boldsymbol{s}^\star_{\text{prob}}\) back into \(F\) yields the closed form
\begin{align}
\min_{\boldsymbol{s}} F(\boldsymbol{s})
&={\small (\boldsymbol{\mu}_1-\boldsymbol{\mu}_2)^\top\,\boldsymbol{\Sigma}_1^{-1}\big(\boldsymbol{\Sigma}_1^{-1}+\boldsymbol{\Sigma}_2^{-1}\big)^{-1}\boldsymbol{\Sigma}_2^{-1}\,(\boldsymbol{\mu}_1-\boldsymbol{\mu}_2)} \notag\\
&=(\boldsymbol{\mu}_1-\boldsymbol{\mu}_2)^\top\,(\boldsymbol{\Sigma}_1+\boldsymbol{\Sigma}_2)^{-1}\,(\boldsymbol{\mu}_1-\boldsymbol{\mu}_2),
\end{align}
using the equation \(A(A+B)^{-1}B=(A^{-1}+B^{-1})^{-1}\). By the matrix AM–HM inequality,
\begin{equation}
(\boldsymbol{\Sigma}_1+\boldsymbol{\Sigma}_2)^{-1}\ \preceq\ \frac{1}{2}\big(\boldsymbol{\Sigma}_1^{-1}+\boldsymbol{\Sigma}_2^{-1}\big),
\end{equation}
we have
\begin{equation}
    \min_{\boldsymbol{s}} F(\boldsymbol{s})\ \le\ \ell_{\mathrm{DCL}}^{(\boldsymbol{\mu})}.
\end{equation}
Thus, the DCL mean term upper-bounds, and therefore drives down, the auxiliary WLS objective whose minimizer is the precision-weighted centroid \(\boldsymbol{s}^\star\). When the across-view noise that perturbs the means satisfies a calibration \(\boldsymbol{\Omega}_i\approx \boldsymbol{\Sigma}_i\) (or \(\boldsymbol{\Omega}_i=\alpha_i \boldsymbol{\Sigma}_i\) with bounded \(\alpha_i>0\))\footnote{This is consistent with a broad line of probabilistic embedding works where each sample is modeled by a Gaussian in the embedding space, with the mean encoding the representation and the covariance reflecting view-induced ambiguity, without claiming identifiability of the true perturbation covariance. For example,~\cite{kirchhof2023probabilistic} explicitly learns a posterior distribution in latent space and connects it to aleatoric uncertainty recovery under modeling assumptions, reinforcing that ``posterior covariance as uncertainty'' is a reasonable and commonly adopted modeling stance.}, the same precision weights are optimal for estimating \(\boldsymbol{s}\) in the sense of the weighted least squares; in particular, the corresponding estimator $\boldsymbol{s}^\star_{\text{prob}}$ has error covariance \((\boldsymbol{\Omega}_1^{-1}+\boldsymbol{\Omega}_2^{-1})^{-1}\), which is no larger than \(\frac{1}{4}(\boldsymbol{\Omega}_1+\boldsymbol{\Omega}_2)\). We prove in the following theorem.
\begin{theorem}[ProbCL strictly improves estimation risk]
\label{thm:hetero}
Under the setup above, the ProbCL precision-weighted estimator has no larger error covariance than the deterministic baseline in the Loewner order:
\begin{equation}
\big(\boldsymbol{\Omega}_1^{-1}+\boldsymbol{\Omega}_2^{-1}\big)^{-1} \;\preceq\; \frac{1}{4}(\boldsymbol{\Omega}_1+\boldsymbol{\Omega}_2).
\end{equation}
Equivalently, for any vector $\boldsymbol{u}\in\mathbb R^d$,
\begin{equation}
\boldsymbol{u}^\top\mathrm{Cov}(\hat{\boldsymbol{s}}_{\text{prob}})\boldsymbol{u} \;\le\; \boldsymbol{u}^\top\mathrm{Cov}(\hat{\boldsymbol{s}}_{\text{det}})\boldsymbol{u}.
\end{equation}
In particular, the mean-squared error (MSE) and any monotone matrix functional (e.g., trace) are no larger for ProbCL than for the deterministic baseline, with equality if and only if $\boldsymbol{\Omega}_1=\boldsymbol{\Omega}_2$.
\end{theorem}

\begin{proof}
\emph{Scalar-wise case.}
Fix a coordinate with variances $\sigma_1^2,\sigma_2^2>0$.
Then
\begin{equation}
\operatorname{Var}(\hat s_{\text{prob}})=\frac{1}{\frac{1}{\sigma_1^2}+\frac{1}{\sigma_2^2}}
=\frac{\sigma_1^2\sigma_2^2}{\sigma_1^2+\sigma_2^2},
\quad
\operatorname{Var}(\hat s_{\text{det}})=\frac{\sigma_1^2+\sigma_2^2}{4}.
\end{equation}
By AM–GM,
$4\sigma_1^2\sigma_2^2 \le (\sigma_1^2+\sigma_2^2)^2$,
hence
\begin{equation}
\frac{\sigma_1^2\sigma_2^2}{\sigma_1^2+\sigma_2^2} \;\le\; \frac{\sigma_1^2+\sigma_2^2}{4},
\end{equation}
with equality iff $\sigma_1^2=\sigma_2^2$.

\emph{Matrix (diagonal/block-diagonal) case.}
Apply the scalar inequality coordinate-wise to each block; stacking the coordinates yields
\begin{equation}
\big(\boldsymbol{\Sigma}_1^{-1}+\boldsymbol{\Sigma}_2^{-1}\big)^{-1} \;\preceq\; \frac{1}{4}(\boldsymbol{\Sigma}_1+\boldsymbol{\Sigma}_2),
\end{equation}
which establishes the Loewner-order claim.
\end{proof}

Note that even without exact calibration, the variance-alignment term encourages \(\boldsymbol{\Sigma}_i\) to reflect view reliability, so the precision weighting remains effective and explains the observed gains in downstream performance. This theoretical expectation is empirically validated in Sec.~\ref{sec: experiment} through our augmentation-strength and downstream predictive-entropy analysis.

In conclusion, a smaller estimation error means that the learned representation stays closer to the underlying augmentation-invariant content of each image and is less contaminated by view-specific artifacts, thereby explicitly mitigating spurious invariance. By down-weighting unreliable views through learned precisions, the probabilistic contrastive learning objective preserves task-relevant variability instead of collapsing it, yielding features that are more faithful to the signal and that generalize better in downstream tasks.

\subsection{The Unified Probabilistic SSL Framework}
\label{sec: 3D}

While the aforementioned probabilistic contrastive learning forms the core of the framework, auxiliary SSL pretext tasks can be further incorporated to enrich the representation learning process. These tasks act as additional inductive biases that encourage the model to encode more diverse features.

The proposed probabilistic contrastive learning method is grounded in the principle of ELBO, replacing the reconstruction term with a contrastive learning objective to promote the learning of representations that are invariant to augmentations. By modeling latent embeddings probabilistically, the contrastive objective can naturally extend back to the reconstruction objective, as originally intended. In addition, a predictive objective can also be incorporated. This elegantly aligns with the three primary SSL approaches discussed earlier, integrating them into a unified probabilistic framework. Together, these auxiliary tasks synergistically ensure that the learned representations remain invariant to augmentations while capturing structural and contextual details.

To facilitate generative and predictive learning objectives, we still resort to the reparameterization trick to obtain the samples. After obtaining the samples, we attach them with respective heads, as shown in the following:

\subsubsection{Probabilistic Generative Learning} 

In this pretext task, random patches of the input image are manually masked in the augmentation step, and the model aim to learn to reconstruct these masked regions, thereby capturing fine-grained local details and contextural dependencies. 

Specifically, given a masked input $\boldsymbol{x}_m$ (e.g., $\boldsymbol{x}_1$ or $\boldsymbol{x}_2$ in Sec.~\ref{sec: 3B}) and its ground truth $\boldsymbol{x}$, the encoder generates the probabilistic embedding $q(\boldsymbol{z}_m|\boldsymbol{x}_m)\sim\mathcal{N}(\boldsymbol{z}_m|\boldsymbol{\mu}_m,\boldsymbol{\sigma}_m^2)$, where we can sample $\hat{\boldsymbol{z}}_m$ using reparameterization trick as in Eq.~\ref{eq: reparam}. After that, $\hat{\boldsymbol{z}}_m$ is passed through a reconstruction head $h_r$ to reconstruct the missing regions, producing an output $\hat{\boldsymbol{x}}=h_r(\hat{\boldsymbol{z}}_m)$ of the same size as the original input. The reconstruction head consists of a linear layer that projects the sampled vector $\hat{\boldsymbol{z}}_m$ into a tensor with the shape of the lowest-resolution hierarchical features of SwinUNETR, and a series of transpose convolution layers with upsampling, to output the reconstruction prediction $\hat{\boldsymbol{x}}$. The probabilistic generative learning (ProbGL) loss is given by:
\begin{equation}
    \ell_{\text{ProbGL}} = \| \hat{\boldsymbol{x}} - \boldsymbol{x} \|_1.
\end{equation}

\subsubsection{Probabilistic Predictive Learning} 

Here we consider rotation prediction as the pretext task of predictive learning, which requires the model to understand global orientation changes, further enriching the learned embeddings with contextual and structural information.

Specifically, for the original input $\boldsymbol{x}$, a randomly rotated version $\boldsymbol{x}_r$ is generated during the augmentation step, accompanied by a rotation label $y_r \in \{0^\circ, 90^\circ, 180^\circ, 270^\circ\}$. Then the encoder produces a probabilistic embedding $q(\boldsymbol{z}_r|\boldsymbol{x}_r)$, from which a sample $\hat{\boldsymbol{z}}_r$ is drawn using the reparameterization trick in Eq.~\ref{eq: reparam}. After that, $\hat{\boldsymbol{z}}_r$ is passed through a classification head $h_c$, which is a linear classification layer, outputs a probability distribution $\hat{y}_r=h_c(\hat{\boldsymbol{z}}_r)$ over the rotation classes to predict the rotation label . The proposed probabilistic predictive learning (ProbPL) loss is defined using cross-entropy:
\begin{equation}
    \ell_{\text{ProbPL}} =  - \sum_{r} y_r \log \hat{y}_r.
\end{equation} 

\subsubsection{Final Unified Probabilistic SSL Objective}

Combining the probabilistic contrastive learning (ProbCL) loss, probabilistic generative learning (ProbGL) loss, and probabilistic predictive learning (ProbPL) loss, our final loss is the following unified probabilistic SSL (ProbSSL) objective:
\begin{equation}
    \ell_{\text{ProbSSL}} = \alpha \ell_\text{ProbCL} + \beta \ell_{\text{ProbGL}} + \gamma \ell_{\text{ProbPL}},
    \label{eq: probssl}
\end{equation}
where $\alpha, \beta$ and $\gamma$ are weighting coefficients for the respective loss terms. We show our pseudocode in Algorithm~\ref{alg}.

While it is true that our unification is reflected in the joint loss function, the interaction between contrastive, generative, and predictive paradigms arises through the shared probabilistic latent space. Specifically, all three tasks operate on latent variables drawn from the same variational distribution $q(z|x)$, enabling each task to regularize and shape the latent space in a complementary manner:

\begin{compactitem}
    \item The contrastive loss enforces global semantic consistency.
    \item The generative loss captures fine-grained structural details.
    \item The predictive loss adds supervision on geometric/contextual transformations.
\end{compactitem}

Together, they guide the latent space toward being both discriminative and expressive. Nevertheless, we would like to emphasize that our novelty is not in unifying contrastive, predictive, and generative paradigms. The main novelty still lies in the probabilistic contrastive learning model (Sec~\ref{sec: 3B} and \ref{sec: 3C}), and the predictive and generative paradigms are included as a decoration to make the framework comprehensive and boost the performance.

The integration of these tasks in a probabilistic framework ensures that the learned representations are not only robust to augmentations but also capture meaningful semantic and structural information essential for medical imaging tasks. In the next section, we evaluate the proposed framework across multiple downstream applications, demonstrating its effectiveness and versatility.

\begin{algorithm}[!t]
\caption{Pseudocode of ProbSSL Framework (in PyTorch)}
\label{alg}
\begin{lstlisting}[language=Python]
# $\textcolor{commentcolor}{f}$: backbone (the encoder of SwinUNETR)
# $\textcolor{commentcolor}{g}$: projection head (dual branch 3-layer MLP)
# $\textcolor{commentcolor}{h}$: prediction head (2-layer MLP)
# $\textcolor{commentcolor}{h_r}$: reconstruction head (5 modules of transpose convolution with upsampling)
# $\textcolor{commentcolor}{h_c}$: classification head (1 linear layer)
for $\boldsymbol{x}$ in loader:  
  # $\textcolor{commentcolor}{\texttt{load a minibatch with n samples}}$
  $\boldsymbol{x}_1, \boldsymbol{x}_2 = \tau_1(\boldsymbol{x}), \tau_2(\boldsymbol{x})$  # $\textcolor{commentcolor}{\texttt{augmented views}}$
  $\boldsymbol{h}_1, \boldsymbol{h}_2 = f(\boldsymbol{x}_1), f(\boldsymbol{x}_2)$  # $\textcolor{commentcolor}{\texttt{representations}}$  
  $\boldsymbol{z}_1 \sim \mathcal{N}(\boldsymbol{\mu}_1, \boldsymbol{\sigma}_1^2)$ where $\boldsymbol{\mu}_1, \boldsymbol{\sigma}_1^2 = g(\boldsymbol{h}_1)$  # $\textcolor{commentcolor}{\texttt{embedding}}$
  $\boldsymbol{z}_2 \sim \mathcal{N}(\boldsymbol{\mu}_2, \boldsymbol{\sigma}_2^2)$ where $\boldsymbol{\mu}_2, \boldsymbol{\sigma}_2^2 = g(\boldsymbol{h}_2)$  # $\textcolor{commentcolor}{\texttt{embedding}}$
  sampling $\hat{\boldsymbol{z}}_1, \hat{\boldsymbol{z}}_2$  # $\textcolor{commentcolor}{\texttt{reparameterization trick}}$
  $\hat{\boldsymbol{p}}_1, \hat{\boldsymbol{p}}_2 = h(\hat{\boldsymbol{z}}_1), h(\hat{\boldsymbol{z}}_2)$ # $\textcolor{commentcolor}{\texttt{predictions of embeddings}}$
  $\hat{\boldsymbol{x}}_1, \hat{\boldsymbol{x}}_2 = h_r(\hat{\boldsymbol{z}}_1), h_r(\hat{\boldsymbol{z}}_2)$ # $\textcolor{commentcolor}{\texttt{reconstructions}}$
  $\hat{y}^r_1, \hat{y}^r_2 = h_c(\hat{\boldsymbol{z}}_1), h_c(\hat{\boldsymbol{z}}_2)$ # $\textcolor{commentcolor}{\texttt{rotation predictions}}$ 
  $\ell_\text{ProbCL} \leftarrow \boldsymbol{\mu}_1, \boldsymbol{\sigma}_1^2, \boldsymbol{\mu}_2, \boldsymbol{\sigma}_2^2, \hat{\boldsymbol{z}}_1, \hat{\boldsymbol{z}}_2$ # $\textcolor{commentcolor}{\texttt{ProbCL loss}}$
  $\ell_\text{ProbGL} \leftarrow \hat{\boldsymbol{x}}_1, \hat{\boldsymbol{x}}_2, \boldsymbol{x}_1, \boldsymbol{x}_2$ # $\textcolor{commentcolor}{\texttt{ProbGL loss}}$
  $\ell_\text{ProbPL} \leftarrow \hat{y}^r_1, \hat{y}^r_2, y^r_1, y^r_2$ # $\textcolor{commentcolor}{\texttt{ProbPL loss}}$
  $\ell_\text{ProbSSL} = \alpha \ell_\text{ProbCL} + \beta \ell_{\text{ProbGL}} + \gamma \ell_{\text{ProbPL}}$ # $\textcolor{commentcolor}{\texttt{loss}}$
  $\ell.\text{backward}()$  # $\textcolor{commentcolor}{\texttt{back-propagation}}$
  $\text{update}(f, g, h, h_r, h_c)$  # $\textcolor{commentcolor}{\texttt{SGD update}}$
\end{lstlisting}
\end{algorithm}

\section{Experiments}
\label{sec: experiment}

In this section, we conduct experiments to evaluate the performance of the proposed probabilistic self-supervised learning framework. The goal of the experiments is to assess the ability of the model to learn useful representations without relying on extensive labeled data and to perform well on downstream tasks after fine-tuning. 

\subsection{Datasets}

We use a collection of unlabeled CT scans in the pre-training phase and several labeled datasets in the fine-tuning phase. In Table~\ref{table: datasets}, we provide an overview of the datasets in our experiments, which are all public.  The details are as follows.











\subsubsection{Pre-training Datasets}

To make a fair comparison with existing methods, we follow the setting of the recent state-of-the-art medical SSL method~\cite{wu2024voco} that uses the combination of the following datasets to pre-train all methods: 
\begin{itemize}
    \item BTCV~\cite{landman2015miccai} with multi-organ abdominal CTs;
    \item TCIA-Covid19~\cite{desai2020chest} with chest CTs from Covid patients;
    \item LUNA16~\cite{setio2017validation} with lung CTs for lung nodule detection.
\end{itemize}

In total, 1,589 CT scans are used for training and 100 for validation. It is important to note that all annotations of these datasets are excluded during the pre-training phase.

\subsubsection{Fine-tuning Datasets} 

After pre-training, we fine-tune with the labeled data from the downstream tasks. To evaluate the versatility of the learned representations, we select the following four different downstream tasks (the data split exactly follows the prior studies~\cite{tang2022self,wu2024voco} for fair comparison):
\begin{itemize}
    \item BTCV~\cite{landman2015miccai}: we fine-tune using its labeled data for the task of multi-organ segmentation. Following~\cite{chen2023masked,wu2024voco}, we only use BTCV and TCIA-Covid19 as the pre-training dataset for fair comparison. The familiarity with its CT distribution during pre-training may reduce the adaptation difficulty. (\ref{sec: downstream1})
    \item MM-WHS~\cite{zhuang2018multivariate}, with the task of segmenting seven whole heart substructures for cardiac CTs. It is more challenging than the first task, as the CT scans are unseen during pre-training and the model is required to generalize to new cardiac structures with limited supervision. (\ref{sec: downstream2}) 
    \item CC-CCII~\cite{zhang2020clinically}, a CT dataset for COVID-19 classification with three classes: novel coronavirus pneumonia (NCP), common pneumonia (CP), and normal controls (Normal). Shifting the task from segmentation to classification, we evaluate the versatility of the learned model. (\ref{sec: downstream3})
    \item BraTS~\cite{baid2021rsna}, an MRI dataset for brain tumor segmentation with three classes: whole tumor (WT), tumor core (TC), and enhancing tumor (ET). Each image consists of four channels, FLAIR, T1w, T1gd and T2w. It escalates challenge by shifting from CT to MRI modality, evaluating the cross-modal generalization ability. (\ref{sec: downstream4})
\end{itemize}

\begin{table}[t!]
\begin{center}
\caption{The datasets used in our experiments (PT for pre-train, FT for fine-tune, Seg. for segmentation, Cls. for classification)}
\label{table: datasets}
{\scriptsize
\begin{tabular}{ccccccc}
\toprule 
Dataset & Modality & Phase & Task & Class & Train/Valid \\
\midrule
BTCV~\cite{landman2015miccai} & CT & PT & -- & -- & 24/6 \\
TCIA Covid19~\cite{desai2020chest}  & CT & PT & -- & -- & 722/49 \\
LUNA~\cite{setio2017validation} & CT & PT & -- & -- & 843/45\\
\midrule
BTCV~\cite{landman2015miccai} & CT & FT & Seg. & 13 & 24/6 \\
MM-WHS~\cite{zhuang2018multivariate} & CT & FT & Seg. & 7 & 14/6 \\
CC-CCII~\cite{zhang2020clinically} & CT & FT & Cls. & 3 & 2514/1664 \\ 
BraTS~\cite{baid2021rsna} & MRI & FT & Seg. & 3 & 387/97 \\ 
\bottomrule
\end{tabular}
}
\end{center}
\end{table}

\subsection{Experimental Setup}

\begin{table*}[!htbp]
\begin{center}
\caption{Experimental results on the BTCV abdominal multi-organ segmentation dataset using Dice scores. The best result is shown in bold. Spl: Spleen, Rkid: Right Kidney, LKid: Left Kidney, Gall: Gallbladder, Eso: Esophagus, Liv: Liver, Sto: Stomach, Aor: Aorta, IVC: Inferior Vena Cava, Veins: Portal and Splenic Veins, Pan: Pancreas, RAG: Right Adrenal Gland, LAG: Left Adrenal Gland, Avg: Average.}
\label{table: btcv}
\resizebox{\textwidth}{!}{
\begin{tabular}{l cccccccccccccc}
\toprule
\textbf{Method} & \textbf{Spl} & \textbf{RKid} & \textbf{LKid} & \textbf{Gall} & \textbf{Eso} & \textbf{Liv} & \textbf{Sto} & \textbf{Aor} & \textbf{IVC} & \textbf{Veins} & \textbf{Pan} & \textbf{RAG} & \textbf{LAG} & \textbf{Avg} \\
\midrule
\emph{Supervised baselines} & \\
~ UNETR~\cite{hatamizadeh2022unetr} & 93.02 & 94.13 & 94.12 & 66.99 & 70.87 & 96.11 & 77.27 & 89.22 & 82.10 & 70.16 & 76.65 & 65.32 & 59.21 & 79.82 \\
~ SwinUNETR~\cite{hatamizadeh2021swin} & 94.06 & 93.54 & 93.80 & 65.51 & 74.60 & 97.09 & 75.94 & 91.80 & 82.36 & 73.63 & 75.19 & 68.00 & 61.11 & 80.53 \\
\midrule
\emph{General SSL methods} & \\
~ SimCLR~\cite{chen2020simple} & 92.79 & 93.04 & 91.41 & 49.65 & 50.99 & 98.49 & 77.92 & 85.56 & 80.58 & 64.37 & 67.16 & 59.04 & 48.99 & 73.85 \\
~ MoCov3~\cite{he2020momentum,chen2021empirical} &  91.96 & 92.85 & 92.42 & 68.25 & 72.77 & 94.91 & 78.82 & 88.21 & 81.59 & 71.15 & 75.76 & 66.48 & 58.81 & 79.54 \\
~ SimMIM~\cite{xie2022simmim}$^\dagger$ & 95.56 & 95.82 & 94.14 & 52.06 & 53.52 & 98.98 & 80.25 & 88.11 & 82.98 & 66.49 & 69.16 & 60.88 & 50.45 & 76.03 \\
~ MAE3D~\cite{he2022masked,chen2023masked} & 93.98 & 94.37 & 94.18 & 69.86 & 74.65 & 96.66 & 80.40 & 90.30 & 83.13 & 72.65 & 77.11 & 67.34 & 60.54 & 81.33 \\
\midrule 
\emph{Medical SSL methods} \\
~ ModelsGenesis~\cite{zhou2021models} & 91.99 & 93.52 & 91.81 & 65.11 & 76.14 & 95.98 & 86.88 & 89.29 & 83.59 & 71.79 & 81.62 & 67.97 & 63.18 & 81.45 \\
~ ROT~\cite{taleb20203d} & 91.75 & 93.13 & 91.62 & 65.09 & 76.55 & 94.21 & 86.16 & 89.74 & 83.08 & 71.13 & 81.55 & 67.90 & 63.72 & 81.20 \\
~ Rubik++~\cite{tao2020revisiting} & 96.21 & 90.41 & 89.33 & 75.22 & 72.64 & 97.44 & 79.25 & 89.65 & 83.76 & 74.74 & 78.35 & 67.14 & 61.97 & 81.38 \\
~ PCRLv1~\cite{zhou2021preservational} & 95.73 & 89.66 & 88.53 & 75.41 & 72.33 & 96.20 & 78.99 & 89.11 & 83.06 & 74.47 & 77.88 & 67.02 & 61.85 & 80.78 \\
~ PCRLv2~\cite{zhou2023unified} & 95.50 & 91.43 & 89.52 & 76.15 & 73.54 & 97.28 & 79.64 & 90.16 & 84.17 & 75.20 & 78.71 & 68.74 & 62.93 & 81.74 \\
~ SwinUNETR~\cite{tang2022self} & 95.25 & 93.16 & 92.97 & 63.62 & 73.96 & 96.21 & 79.32 & 89.98 & 83.19 & 76.11 & 82.25 & 68.99 & 65.11 & 81.54 \\
~ SwinMM~\cite{wang2023swinmm} & 94.33 & 94.18 & 94.16 & 72.97 & 74.75 & 96.37 & 83.23 & 89.56 & 82.91 & 70.65 & 75.52 & 69.17 & 62.90 & 81.81 \\
~ GVSL~\cite{he2023geometric} & 95.27 & 91.22 & 92.25 & 72.69 & 73.56 & 96.44 & 82.40 & 88.90 & 84.22 & 70.84 & 76.42 & 67.48 & 63.25 & 81.87 \\
~ GL-MAE~\cite{zhuang2025advancing} & 95.59 & 94.39 & 94.37 & 75.79 & 75.93 & 97.91 & 86.99 & 90.34 & 83.11 & 70.80 & 75.71 & 69.89 & 69.12 & 83.10 \\
~ VoCo~\cite{wu2024voco} & 95.73 & 96.53 & 94.48 & 76.02 & 75.60 & 97.41 & 78.43 & 91.21 & 86.12 & 78.19 & 80.88 & 71.47 & 67.88 & 83.85 \\
~ MiM~\cite{zhuang2025mim} & 96.05 & 94.58 & 94.53 & 75.73 & 77.36 & 97.03 & 83.23 & 90.37 & 87.64 & 74.97 & 82.62 & 71.02 & 68.80 & 84.46 \\
~ S2DC~\cite{pan2025structure} & 96.24 & 96.51 & 94.52 & 74.27 & 77.69 & 96.95 & 83.33 & 90.44 & 86.66 & 75.22 & 84.24 & 68.22 & 69.73 & 84.14 \\
~ \textbf{ProbSSL (Ours)} & 96.02 & 95.78 & 94.51 & 78.54 & 78.28 & 96.91 & 85.75 & 91.06 & 86.12 & 76.69 & 82.90 & 71.30 & 69.74 & \textbf{84.89} \\
\bottomrule
\end{tabular}
}
\begin{minipage}{0.98\textwidth}
\footnotesize
Note: The results of comparison methods are reported in~\cite{wu2024voco,zhuang2025advancing,zhuang2025mim,pan2025structure}. All methods are evaluated using the \textit{same pre-training datasets, fine-tuning data splits (train/validation/test), and evaluation protocols} to ensure a fair and consistent comparison.
\end{minipage}
\end{center}
\end{table*}

\subsubsection{Implementation Details}

We use two NVIDIA A100 GPUs, each with 40 GB of memory, for pre-training and a single GPU for fine-tuning on downstream tasks. Our method is implemented using the PyTorch and MONAI~\cite{cardoso2022monai} libraries.

During pre-training, CT scans undergo intensity normalization by standardizing voxel intensities within the clipped range $[-1000, 1000]$ and spatial cropping to extract input patches of size $96\times 96\times 96$. For data augmentation, we apply random rotation and random region masking to generate positive pairs for probabilistic contrastive learning, while ensuring compatibility with the other two probabilistic SSL pretext tasks. The initial learning rate is 6e-6 and the batch size is 1 per GPU. We pre-train the model for 100,000 iterations (approximately 125 epochs). For the optimizer, we use AdamW~\cite{loshchilov2018decoupled} with a learning rate schedule consisting of a 500-iteration linear warm-up followed by cosine decay. For the loss weights in Eq.~\ref{eq: probcl}, the hyperparameters $\alpha_1, \alpha_2, \alpha_3$, which correspond to distinct components of the probabilistic contrastive learning loss, are set to 1, 1, and 0.1, respectively, to balance the scale of the losses. The loss weights $\alpha, \beta, \gamma$ are assigned equally as 1 as they are roughly on the same scale, following prior multi-task SSL method~\cite{tang2022self}.

During fine-tuning, we also use the backbone of SwinUNETR~\cite{hatamizadeh2021swin} for segmentation tasks. For classification task, we adopt SwinViT as the backbone instead, maintaining the same encoder structure as SwinUNETR while replacing the decoder with a linear classification layer.

\subsubsection{Comparison Methods}

We compare our method with several state-of-the-art self-supervised learning methods, which are tailored to both general vision tasks and medical-specific tasks. For general SSL methods, we compare with SimCLR~\cite{chen2020simple}, MoCo~\cite{he2020momentum}, MAE~\cite{he2022masked} and SimMIM~\cite{xie2022simmim}, where SimCLR and MoCo represent contrastive learning methods, while MAE and SimMIM are based on generative learning. Since these methods were not originally designed for medical tasks, we adapt their parameters to better align with medical training, following \cite{chen2023masked,wu2024voco}. For medical SSL methods, we compare with Models Genesis~\cite{zhou2021models}, Rubik++~\cite{tao2020revisiting}, PCRLv1~\cite{zhou2021preservational}, PCRLv2~\cite{zhou2023unified}, SwinUNETR~\cite{tang2022self}, SwinMM~\cite{wang2023swinmm}, GVSL~\cite{he2023geometric}, VoCo~\cite{wu2024voco}, MiM~\cite{zhuang2025mim}, GL-MAE~\cite{zhuang2025advancing}, and S2DC~\cite{pan2025structure}, all of which have shown effectiveness in 3D medical tasks. We also include supervised baselines trained from scratch.

\subsubsection{Evaluation Metrics} We assess performance using standard metrics for each medical imaging task. For segmentation, we report Dice score (\emph{Dice}) to measure region overlap and Hausdorff 95\% distance (\emph{HD95}) to quantify boundary error. For classification, while the area under the Receiver Operating Characteristic curve (AUROC) is widely used in medical imaging, we adopt accuracy (\emph{Acc.}) in this study for fair comparison with prior work.

\subsection{Comparison with the State-of-the-Arts}
\label{sec: sota}

In the following, we report the experimental results of our ProbSSL method on different downstream medical imaging tasks, compared to state-of-the-art self-supervised learning methods, both general and specific to medical tasks.

\subsubsection{Experimental Results on BTCV Dataset}
\label{sec: downstream1}

First, we evaluate our method on the BTCV dataset, where the training inputs were included in the pre-training phase, but the labels were not. Now we further fine-tune the pre-trained model using the labels and then assess the performance on its validation data.

In Table~\ref{table: btcv}, we report Dice scores computed with a sliding window inference overlap of 0.75, the same setting as the recent state-of-the-art method VoCo~\cite{wu2024voco}. The supervised baselines, UNETR and SwinUNETR, were trained from scratch, achieving the Dice of 79.82 and 80.53. Our ProbSSL, applied on SwinUNETR, significantly outperforms these baselines with Dice 84.89, emphasizing the benefits of self-supervised pre-training. In addition, medical SSL methods generally outperform general SSL methods by taking advantage of domain knowledge. Notably, ProbSSL outperforms all compared general and medical SSL methods, including VoCo which is also based on SwinUNETR architecture, and the recent methods MiM, GL-MAE, and S2DC, all of which follow the same pre-training setup, data split, and evaluation protocols. These results demonstrate the effectiveness of probabilistic feature representations for medical image segmentation.

\begin{table}[!htbp]
\centering
\caption{Experimental results on whole heart segmentation dataset MM-WHS and COVID-19 classification dataset CC-CCII}
\label{tab:mm-whs&cc-ccii}
\begin{tabular}{lccc}
\toprule
\multicolumn{2}{c}{} & \multicolumn{1}{c}{\emph{MM-WHS}} & \multicolumn{1}{c}{\emph{CC-CCII}} \\
\textbf{Method} & \textbf{Backbone} & \textbf{Avg Dice} & \textbf{Avg Acc.}\\
\midrule
\multicolumn{3}{l}{\emph{Supervised baseline}}\\
~ 3D U-Net~\cite{ronneberger2015u} & 3D U-Net & 83.09 & -- \\
~ UNETR~\cite{hatamizadeh2022unetr} & UNETR & 85.85 & -- \\
~ SwinUNETR~\cite{hatamizadeh2021swin} & SwinUNETR & 86.11 & -- \\
~ ResNet~\cite{he2016deep} & ResNet & -- & 85.94  \\
~ ViT~\cite{dosovitskiy2021an} & ViT & -- & 85.92 \\
~ Swin-ViT~\cite{liu2021swin} & SwinViT & -- & 87.15 \\
\midrule
\multicolumn{3}{l}{\emph{General SSL methods}} \\
~ SimCLR~\cite{chen2020simple} & UNETR & 86.00 & 87.12 \\
~ MAE~\cite{he2022masked} & UNETR & 86.03 & 89.47\\
~ MoCo~\cite{he2020momentum} & UNETR & 84.16 & 84.95\\
\midrule
\multicolumn{3}{l}{\emph{Medical SSL methods}}\\
~ Rubik++~\cite{tao2020revisiting} & 3D U-Net & 87.02 & 89.23 \\
~ PCRLv1~\cite{zhou2021preservational} & 3D U-Net & 86.58 & 88.72 \\
~ PCRLv2~\cite{zhou2023unified} & 3D U-Net & 86.82 & 89.15 \\
~ SwinUNETR~\cite{tang2022self} & SwinUNETR & 87.06 & 89.45 \\
~ SwinMM~\cite{wang2023swinmm} & SwinUNETR & 86.98 & 89.61 \\
~ VoCo~\cite{wu2024voco} & SwinUNETR & 90.54 & 90.83 \\
~ \textbf{ProbSSL} & SwinUNETR & \textbf{91.57} & \textbf{92.85} \\
\bottomrule
\end{tabular}
\end{table}

\subsubsection{Experimental Results on MM-WHS Dataset}
\label{sec: downstream2} 

\newlength{\panelht}
\setlength{\panelht}{0.36\textheight}

\begin{figure}[t]
\centering    \includegraphics[width=0.8\linewidth,height=0.2\textheight]{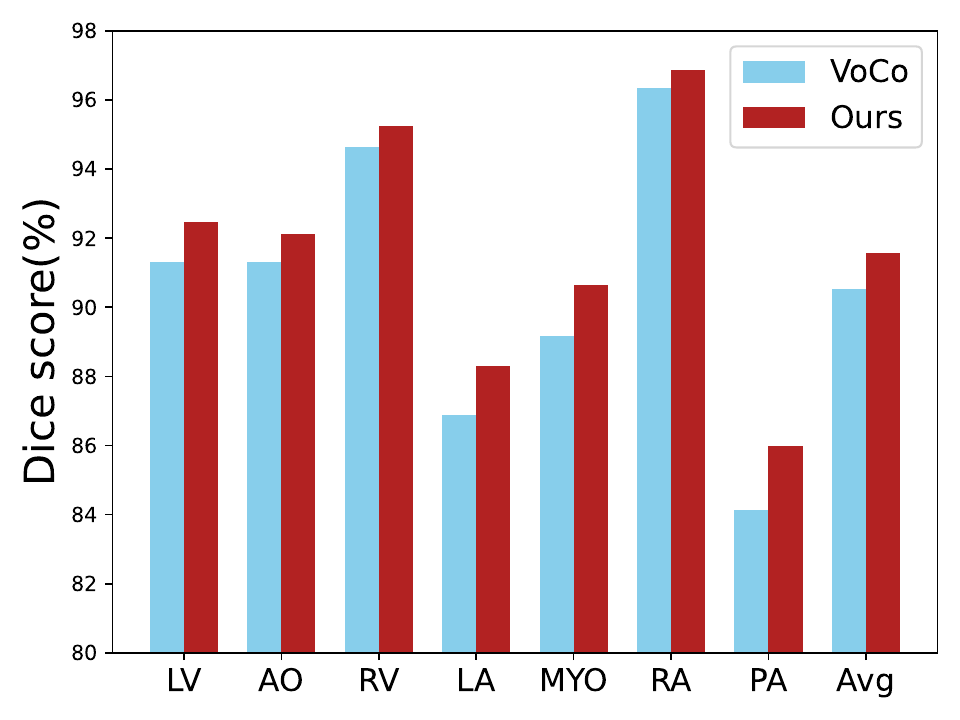}
\caption{Per-organ Dice comparison of our ProbSSL and the state-of-the-art method VoCo on MM-WHS.}
\label{fig:mmwhs-per-organ}
\end{figure}

We then evaluate our method on a downstream MM-WHS dataset that was not used during pre-training. As shown in Table~\ref{tab:mm-whs&cc-ccii}, our proposed ProbSSL method achieves the highest Dice performance among all comparison methods, demonstrating superior generalizability to unseen data. Specifically, ProbSSL outperforms the supervised baseline SwinUNETR by over 5\% and exceeds the state-of-the-art medical SSL method VoCo~\cite{wu2024voco} by over 1\%. We also provide per-organ Dice performance compared with the VoCo in Fig.~\ref{fig:mmwhs-per-organ}. 

We further compare our method against the following three domain experts that are specifically designed for the MM-WHS dataset, all trained using the same amount of data. 
\begin{compactitem}
    \item Two-Stage UNet~\cite{liu2019automatic} presents a two-stage U-Net framework and an adaptive threshold window to automatically segment a whole heart and its substructures.
    \item SRMA-KD~\cite{chen2025srma} introduces a knowledge distillation approach specifically designed for cardiac image segmentation, incorporating three key components: the Global Structural Relational Block (GSRB), the Multi-scale Feature Attention Block (MFAB), and the Prediction Difference Transfer Block (PDTB). These modules significantly enhance segmentation accuracy for cardiac image segmentation compared to other medical imaging knowledge distillation methods.
    \item Hybrid 3D UNet~\cite{kotte2025hybrid} presents a hybrid 3D U-Net framework that utilizes residual connections with attention mechanisms to improve feature learning and localization of cardiac structures.
\end{compactitem}
The results, summarized in Table~\ref{table: mmwhs}, show that our ProbSSL outperforms these domain expert models by a large margin, highlighting the advantages of leveraging self-supervised learning for cardiac substructure segmentation.

\begin{table}[t]
\centering
\caption{Quantitative comparison of segmentation performance on Dice between our ProbSSL and domain experts that are specifically designed for the MM-WHS dataset.}
\label{table: mmwhs}
\resizebox{\columnwidth}{!}{
\begin{tabular}{lccccccccccc}
\toprule
\textbf{Method} & \textbf{MYO} & \textbf{LA} & \textbf{LV} & \textbf{RA} & \textbf{RV} & \textbf{AO} & \textbf{PA} & \textbf{Avg. Dice$\uparrow$} \\
\midrule
Two-Stage UNet~\cite{liu2019automatic} & 75.6 & 82.1 & 86.5 & 66.7 & 77.1 & 75.6 & 72.3 & 79.3 \\
SRMA-KD~\cite{chen2025srma}    & 82.7 & 82.8 & 89.5 & 85.2 & 87.2 & 89.2 & 79.5 & 85.3 \\
Hybrid 3D UNet~\cite{kotte2025hybrid} & 85.3 & 87.6 & 84.9 & 88.4 & 86.1 & 83.4 & 80.5 & 85.8 \\
\midrule
Our ProbSSL & 90.65 & 88.31 & 92.48 & 96.87 & 95.23 & 92.11 & 85.98 & 91.57 \\
\bottomrule
\end{tabular}
}
\end{table}

\begin{figure*}[t]
  \centering
   \includegraphics[width=0.9\linewidth]{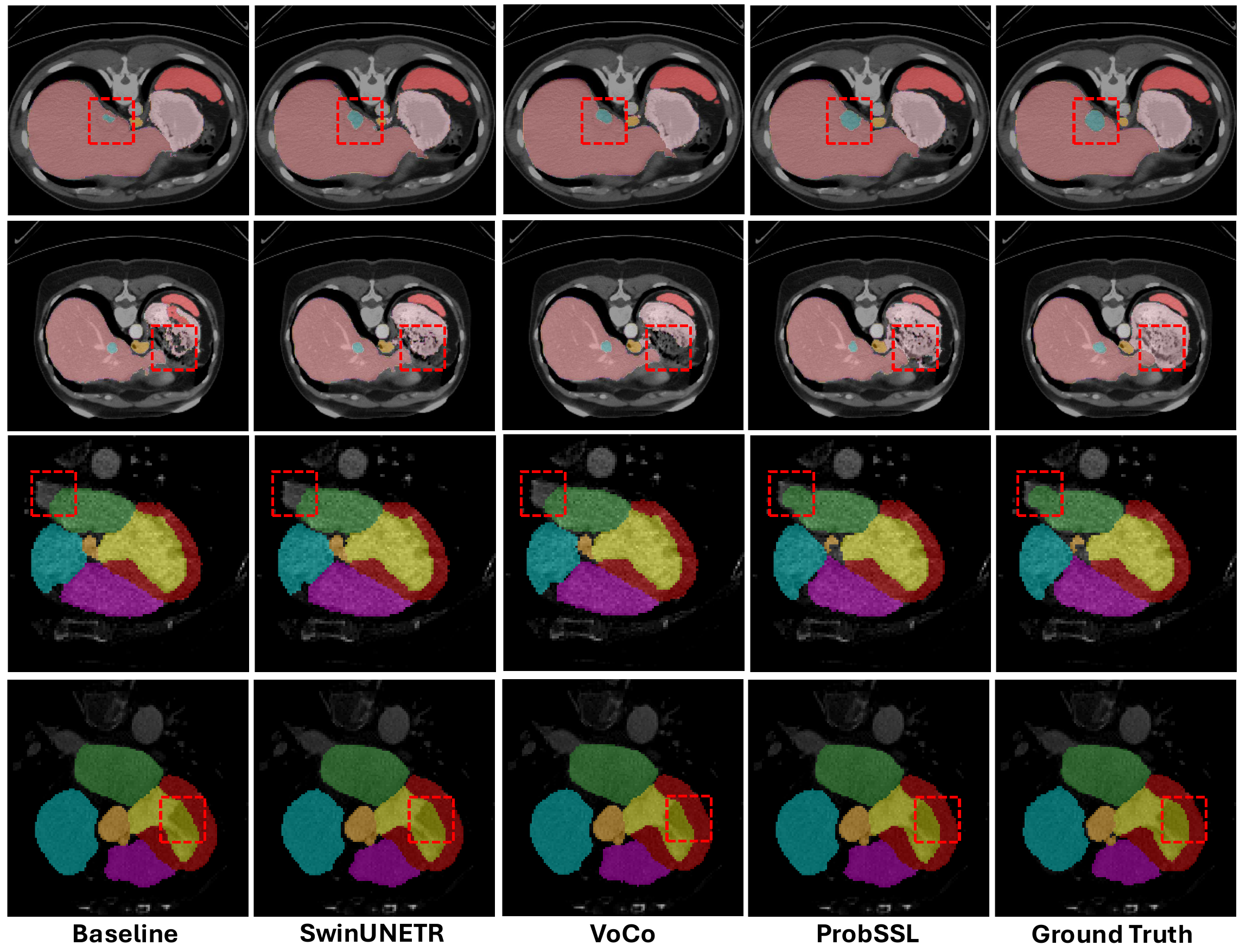}
   \caption{Qualitative visualizations illustrating segmentation performance on test samples from the BTCV dataset (rows 1-2) and MM-WHS dataset (rows 3-4). Differences are highlighted with red boxes (best viewed in color).}
   \label{fig:qualitative}
\end{figure*}

\subsubsection{Experimental Results on CC-CCII Dataset}
\label{sec: downstream3}

Previously, we evaluated our method on segmentation tasks. We now turn to a classification task using the CC-CCII dataset, which is designed for the COVID-19 classification. As shown in Table~\ref{tab:mm-whs&cc-ccii}, our ProbSSL method outperforms both supervised and self-supervised baselines, achieving the highest average accuracy of 92.85\%. Since this is a classification task, we replace the U-shaped architecture like SwinUNETR with standard classification networks—ResNet, ViT, and SwinViT—for the supervised baselines. Among these baselines, SwinViT achieves an accuracy of 87.15\%, less than our ProbSSL by more than 5\%. Compared to SSL methods, our ProbSSL also surpasses the state-of-the-art VoCo by over 2\%. These results highlight the benefits of probabilistic feature representations tailored to medical image data. By effectively modeling the inherent variability in medical images, ProbSSL not only boosts classification accuracy but also demonstrates robust generalization capabilities, underscoring its potential for real-world medical applications.

\begin{table}[t]
\centering
\caption{Experimental results on the brain tumor segmentation dataset BraTS}
\label{tab:brats}
\begin{tabular}{lccccc}
\toprule
\textbf{Method} & \textbf{WT} & \textbf{TC} & \textbf{ET} & \textbf{Avg}\\
\midrule
\multicolumn{5}{l}{\emph{Supervised baseline}}\\
~ UNETR~\cite{hatamizadeh2022unetr} & 81.62 & 87.81 & 57.34 & 75.58\\
~ SwinUNETR~\cite{hatamizadeh2021swin} & 81.28 & 88.67 & 57.73 & 75.89 \\
\midrule
\multicolumn{5}{l}{\emph{General SSL methods}} \\
~ SimCLR~\cite{chen2020simple} & 83.13 & 89.44 & 58.42 & 76.99 \\
~ MoCo~\cite{he2020momentum} & 82.60 & 88.89 & 57.69 & 76.39 \\
~ MAE~\cite{he2022masked} & 82.34 & 90.35 & 59.18 & 77.29 \\
~ SimMIM~\cite{xie2022simmim} & 84.06 & 90.43 & 59.07 & 77.85 \\
\midrule
\multicolumn{5}{l}{\emph{Medical SSL methods}}\\
~ PCRLv1~\cite{zhou2021preservational} & 81.96 & 88.83 & 57.58 & 76.12 \\
~ PCRLv2~\cite{zhou2023unified} & 82.13 & 90.06 & 57.70 & 76.63 \\
~ Rubik++~\cite{tao2020revisiting} & 84.32 & 90.23 & 58.01 & 77.51 \\
~ Swin-UNETR~\cite{tang2022self} & 82.51 & 89.08 & 58.15 & 76.58 \\
~ SwinMM~\cite{wang2023swinmm} & 83.48 & 90.47 & 58.72 & 77.56 \\
~ VoCo~\cite{wu2024voco} & 85.27 & 90.45 & 59.87 & 78.53 \\
~ \textbf{ProbSSL} & 85.29 & 91.29 & 60.58 & \textbf{79.05} \\
\bottomrule
\end{tabular}
\end{table}

\begin{figure*}[t]
  \centering
  \includegraphics[width=0.9\linewidth]{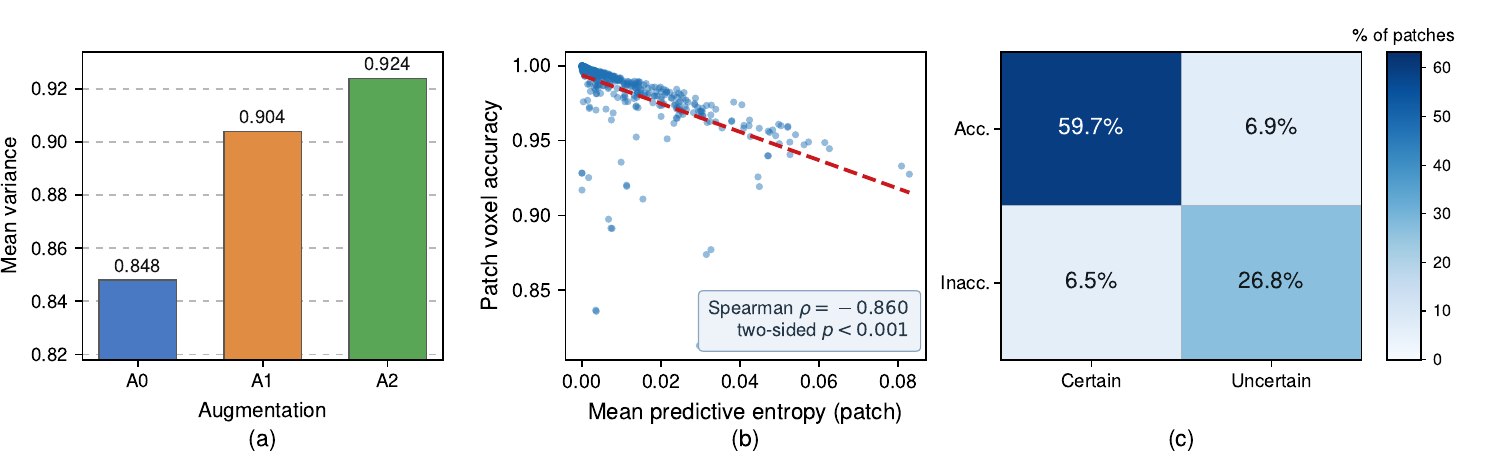}
  \caption{Analyses of augmentation-induced latent variability and downstream predictive entropy. (a) During contrastive pre-training, mean marginal predictive variance increases from A0 to A2 as augmentation strength increases.(b) On BTCV validation, patch-level predictive entropy is negatively correlated with patch accuracy. (c) On BTCV validation, PAvPU-style patch partition shows that most patches fall into the accurate--certain and inaccurate--uncertain groups.}
  \label{fig:rebuttal_combined_pretrain_btcv}
\end{figure*}

\subsubsection{Experimental Results on BraTS Dataset}
\label{sec: downstream4}

Previously, our pre-training and fine-tuning datasets were all CT images. Now we turn to MR images, which increases challenges to evaluate cross-modal generalizability. In Table~\ref{tab:brats}, we present segmentation results on the challenging BraTS dataset using a sliding window inference overlap of 0.75. Our ProbSSL method achieves the highest mean Dice score of 79.05\%, surpassing both the supervised baselines (75.89\%) and the best medical SSL methods VoCo (78.53\%). Notably, ProbSSL demonstrates consistent gains across the three tumor subregions, indicating the robustness of its learned representations. These results underscore the effectiveness of incorporating probabilistic feature representations into self-supervised learning, allowing the model to capture subtle variations in tumor appearance and achieve more accurate segmentation.

\subsection{Qualitative Results and Boundary Error Analysis}
In addition to consistent superior quantitative performance in various downstream tasks, we also show qualitative results of segmentation performance in Fig.~\ref{fig:qualitative}. For BTCV dataset, the supervised Baseline shows significant segmentation omissions, SwinUNETR and VoCo show some improvements, and our ProbSSL provides results closest to the ground truth. For MM-WHS, again, segmentation discrepancies are apparent, especially around the boundaries of the cardiac segments, while our ProbSSL most closely aligns with the ground truth.

To avoid subjectivity, we complement Fig.~\ref{fig:qualitative} with an objective, dataset-level summary of segmentation boundary errors using HD95 (Hausdorff distance at the 95th percentile). Specifically, we compute HD95 per case and report the mean $\pm$ std over the full test set for each method on BTCV and MM-WHS. As shown in Table~\ref{tab:hd95_stats_compact}, our ProbSSL achieves the lowest mean HD95 on both datasets, indicating consistently smaller boundary errors on average.

While masked reconstruction with a pixel-level $L_1$ loss can improve high-frequency details and may partially reduce HD95, this factor alone does not explain our boundary-error gains. Importantly, the SwinUNETR baseline~\cite{tang2022self} already uses the same masked-volume reconstruction objective with an $L_1$ loss, yet ProbSSL still achieves substantially lower HD95 (e.g., $6.63$ vs. $8.59$ on BTCV). Because the reconstruction objective is shared, the additional improvement is more plausibly attributable to the proposed probabilistic latent representation rather than to pixel-level reconstruction alone.

\begin{table}[t]
\centering
\small
\caption{Statistical boundary-error analysis (HD95). Per-case mean HD95 (mm, $\downarrow$) reported as mean$\pm$std over the test set.}
\label{tab:hd95_stats_compact}
{\tiny
\begin{tabular}{lcccc}
\toprule
\textbf{Dataset} & \textbf{Baseline} & \textbf{SwinUNETR} & \textbf{VoCo} & \textbf{ProbSSL} \\
\midrule
\textbf{BTCV}   & $12.63\pm10.61$ & $8.59\pm2.96$ & $8.75\pm3.31$ & $\mathbf{6.63\pm2.42}$ \\
\textbf{MM-WHS} & $5.36\pm2.28$   & $5.12\pm2.09$   & $5.09\pm2.20$    & $\mathbf{4.49\pm1.82}$ \\
\bottomrule
\end{tabular}
}
\end{table}


\subsection{More Analysis of Our Method}

In the final part of this section, we present ablation studies to validate the effectiveness of our proposed method and further analysis to discuss the advantages and limitations of the proposed method. 

\begin{table}[t]
\centering
\caption{BTCV Dice (\%) comparison for contrastive-only pre-training ablations under different augmentation intensities. A0/A1/A2 denote progressively stronger contrastive-view augmentations.}
\label{tab:aug_ablation}
\begin{tabular}{@{}lccc@{}}
\toprule
Method & A0 & A1 & A2 \\ \midrule
Standard CL baseline & 81.07 & 80.75 & 80.65 \\
ProbCL (Ours) & \textbf{83.02} & \textbf{83.20} & \textbf{83.27} \\ \bottomrule
\end{tabular}
\end{table}

\subsubsection{Motivation Validation}
\label{sec:motivation_validation}

To further validate the motivation of our method, we conduct additional analysis to examine whether probabilistic modeling remains beneficial when augmentation-induced ambiguity becomes stronger in medical contrastive learning. Our motivation is that deterministic alignment may enforce overly rigid invariance under aggressive perturbations and thus suppress subtle but task-relevant anatomical cues. To test this directly, we perform a controlled contrastive-only ablation by progressively increasing augmentation strength from A0 to A1 and A2: 
\begin{compactitem}
    \item (\textbf{A0}) rotation + region masking (our default);
    \item (\textbf{A1}) applies A0 and then adds intensity-style medical perturbations (random contrast, additive intensity shift, and additive Gaussian noise), each activated independently with probability $0.4$;
    \item (\textbf{A2}) applies A1 and further stacks a series of 3D affine deformations including anisotropic scaling, shearing, and translation, applied with probability $0.5$.
\end{compactitem}

The results are summarized in Table~\ref{tab:aug_ablation}. We observe that stronger augmentations degrade the deterministic contrastive baseline, which is consistent with recent findings in medical contrastive learning~\cite{idris2025stronger}. In contrast, our ProbCL remains stable and even slightly improves, supporting our central claim that probabilistic alignment mitigates spurious invariance under aggressive medical perturbations.

We further analyze whether the learned variance is designed to capture view-induced ambiguity/variability in the probabilistic embedding space during contrastive pre-training, so that harder augmented views are aligned less aggressively. To verify this claim directly, we ran ProbCL pre-training under three augmentation schedules that differ only in augmentation strength (A0 $<$ A1 $<$ A2), and measured the mean marginal variance, i.e., the average of the diagonal Gaussian variances over training. As shown in Fig.~\ref{fig:rebuttal_combined_pretrain_btcv}(a), this quantity increases monotonically from A0 to A2. Since stronger augmentations are designed to produce harder and more ambiguous views, this monotonic trend shows that the learned variance responds systematically to view difficulty, rather than behaving as a free or arbitrary coefficient in the loss.

To evaluate whether the probabilistic branch learns representations that are meaningfully aligned with downstream prediction reliability, we added an explicit predictive evaluation on BTCV validation. We tiled each validation volume into $96^3$ patches with stride $48$ and retained the $n{=}447$ patches that still contain foreground in the reference mask, so that each patch has a well-defined voxel error rate and a stable patch-level summary. For each retained patch, we computed the mean softmax predictive entropy and the patch voxel accuracy. Here, entropy is used only as a downstream predictive proxy for patch difficulty/reliability; we do not use it interchangeably with the learned latent variance. Rather, these analyses are meant to test whether the probabilistic model yields predictive scores that are aligned with correctness.

As shown in Fig.~\ref{fig:rebuttal_combined_pretrain_btcv}(b), predictive entropy is strongly negatively associated with patch accuracy (Spearman's $\rho=-0.860$, two-sided $p<0.001$). Thus, patches assigned higher predictive entropy are precisely those on which the model is less accurate.

On the same $447$ patches, we further report a $2\times2$ analysis in the spirit of PAvPU~\cite{mukhoti2018evaluating}. We threshold at the arithmetic mean entropy and the arithmetic mean voxel error over the retained patches: a patch is called ``uncertain'' if its mean entropy is at or above the mean entropy, and ``accurate'' if its voxel error is at or below the mean error. As shown in Fig.~\ref{fig:rebuttal_combined_pretrain_btcv}(c), the resulting fractions are accurate--certain $59.7\%$, accurate--uncertain $6.9\%$, inaccurate--certain $6.5\%$, and inaccurate--uncertain $26.8\%$. Therefore, $86.5\%$ of patches fall into the desirable accurate--certain or inaccurate--uncertain groups, indicating that the model's predictive ambiguity is well aligned with patch-level difficulty/error rather than behaving arbitrarily.

Taken together, these analyses provide two complementary pieces of evidence. First, at pre-training time, the learned variance increases systematically with augmentation strength, directly supporting that it tracks view-induced ambiguity and is not merely an arbitrary weight. Second, at downstream evaluation time, predictive entropy is strongly aligned with patch correctness, both through the entropy--accuracy correlation and the PAvPU-style breakdown. 

\subsubsection{Ablation Studies on Our Loss Components}
\label{sec:ablation_loss_components}

\begin{figure}[ht]
  \centering
  \begin{minipage}[t]{0.49\linewidth}
    \centering
    \includegraphics[width=\linewidth]{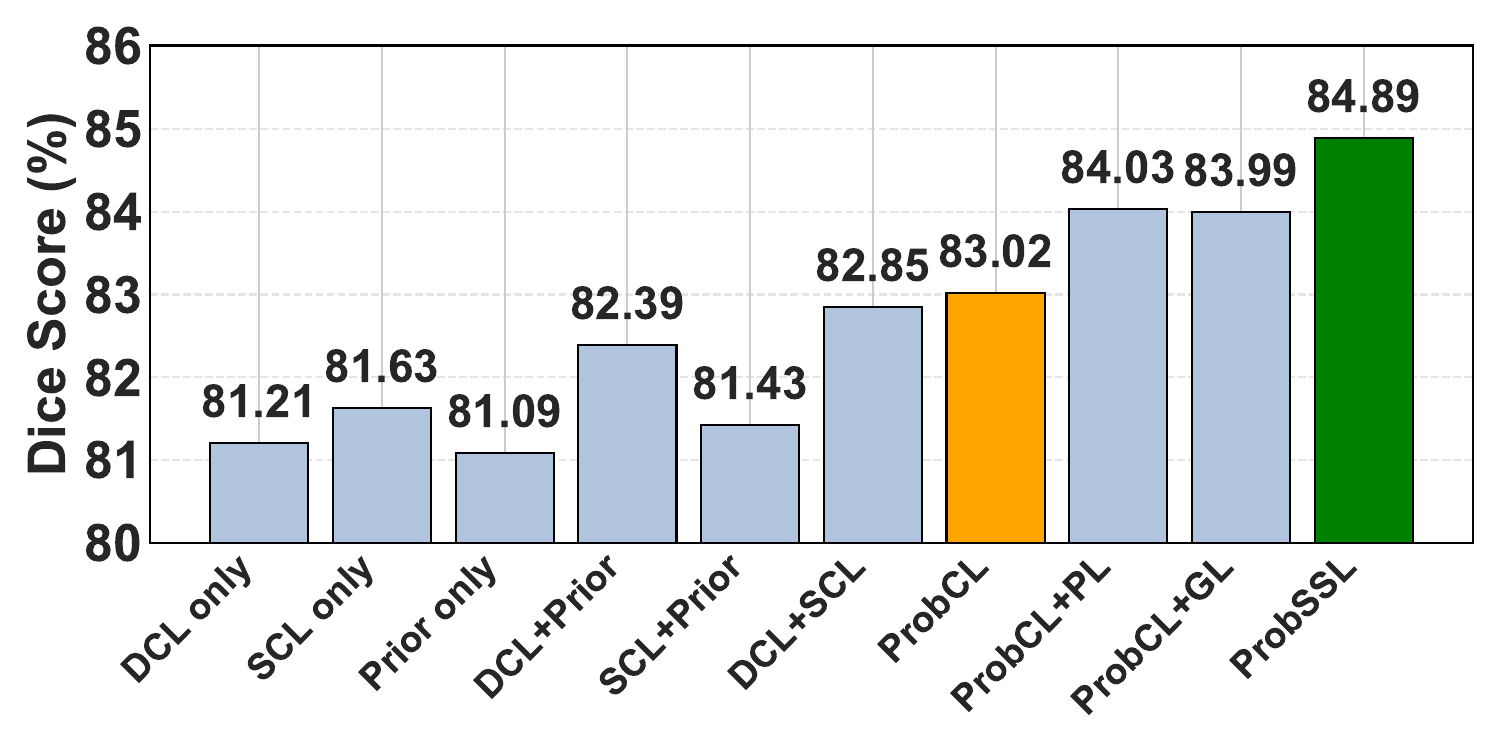}
    {\footnotesize(a) BTCV}
    \label{fig:ablation-1}
  \end{minipage}\hfill
  \begin{minipage}[t]{0.49\linewidth}
    \centering
    \includegraphics[width=\linewidth]{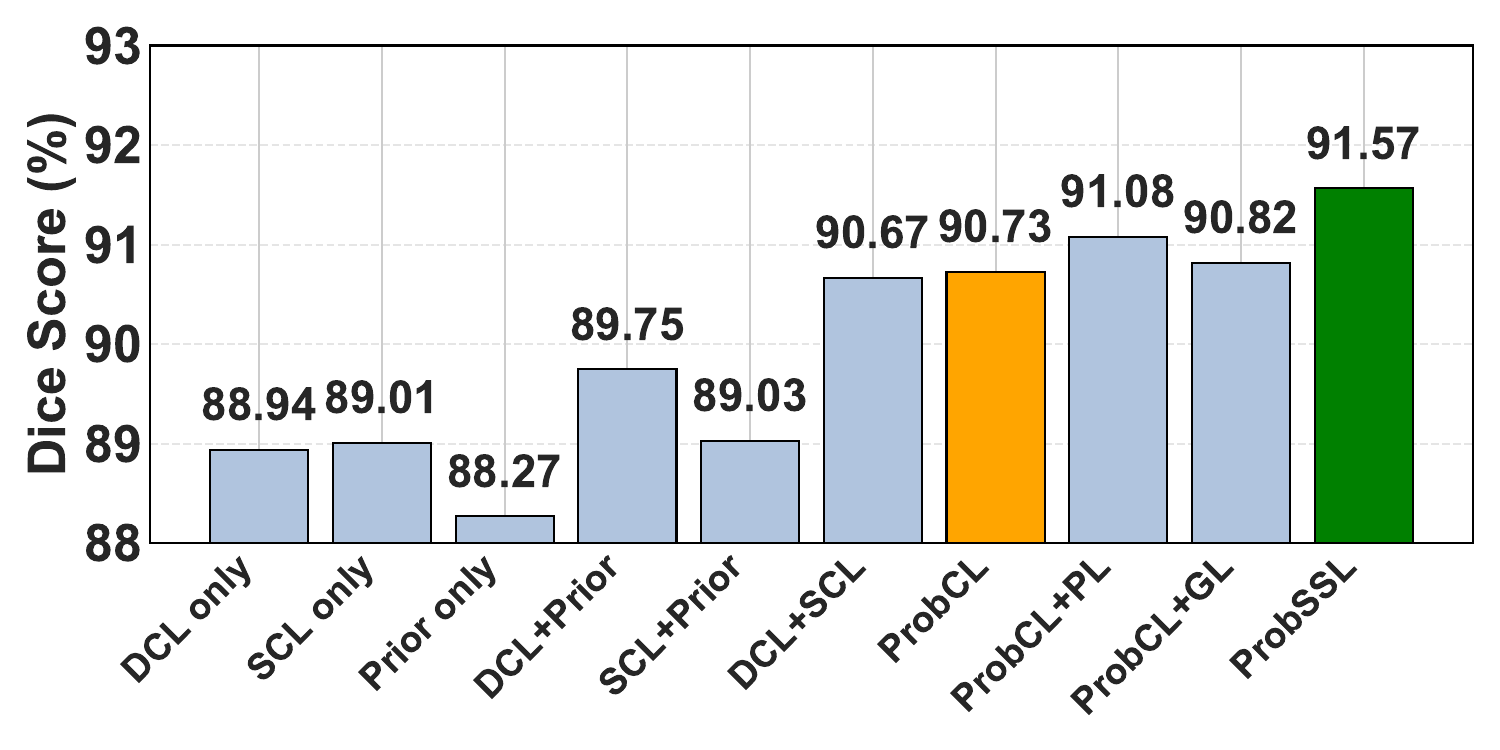}
    {\footnotesize(b) MM-WHS}
    \label{fig:ablation-2}
  \end{minipage}
  \caption{Ablation of loss components on BTCV and MM-WHS. Performance progressively improves as we move from DCL to ProbCL and further to ProbSSL, reflecting the benefit of integrating complementary terms.}
  \label{fig:loss-term-ablation}
\end{figure}

Guided by Eq.~\ref{eq: probcl} and \ref{eq: probssl}, we first conduct a component ablation that toggles the terms inside $\mathcal{L}_{\text{ProbCL}}$ (DCL, SCL, Prior), then incrementally adds the $\mathcal{L}_{\text{ProbGL}}$ and $\mathcal{L}_{\text{ProbPL}}$ branches to form $\mathcal{L}_{\text{ProbSSL}}$. We conduct all ablation studies with the dataset of BTCV and MM-WHS, and the results are shown in Fig.~\ref{fig:loss-term-ablation}.

We begin with the DCL-only variant and progressively enable additional terms inside Eq.~\ref{eq: probcl}. In addition, we explicitly frame the $\mathcal{L}_{\text{DCL}}{=}0$ setting (i.e., removing distribution-level contrast) as a probabilistic contrastive baseline similar to~\cite{duan2024contrastive}, which helps isolate the contribution of our distribution-level alignment.
Within $\mathcal{L}_{\text{ProbCL}}$, the single-term variants DCL and SCL are close, while the prior alone is lowest, reflecting their roles as representation-driving (DCL/SCL) versus regularizing (prior). Most pairs outperform the singles, and the DCL+SCL combination is the strongest pair on both datasets, indicating that distribution-level contrast (DCL) and sample-level contrast (SCL) provide complementary signals. Activating all three weights yields ProbCL, which provides a consistent lift over the best pair, suggesting that the prior helps stabilize the variance estimation. Moving from ProbCL to ProbSSL (Eq.~\ref{eq: probssl}), we further add pretext tasks of prediction (ProbPL) and generation (ProbGL). Combining both add-ons yields the best result on each dataset. In sum, Fig.~\ref{fig:loss-term-ablation} shows that DCL and SCL are complementary representation losses; the prior acts as a variance-regularizing term that modestly improves the full contrastive objective; adding the pretexts of ProbPL and ProbGL yields further gains; and combining all terms (ProbSSL) achieves the highest overall performance.

\begin{table*}[t]
\centering
\caption{Performance comparison based on loss types of three SSL approaches}
\label{tab:ablation_total}
{\scriptsize
\begin{tabular}{lcccccccc}
\toprule
\multirow{2}{*}{Losses} & \multicolumn{2}{c}{BTCV (\emph{in Dice})} & \multicolumn{2}{c}{MM-WHS (\emph{in Dice})} & \multicolumn{2}{c}{CC-CCII (\emph{in Acc.})} & \multicolumn{2}{c}{BraTS (\emph{in Dice})}\\
\cmidrule(lr){2-3} \cmidrule(lr){4-5} \cmidrule(lr){6-7} \cmidrule(lr){8-9}
 & Deterministic & Probabilistic & Deterministic & Probabilistic & Deterministic & Probabilistic & Deterministic & Probabilistic\\
\midrule
Contrastive  & 81.07 & \textbf{83.02} & 87.53 & \textbf{90.73} & 88.96  & \textbf{92.19} & 75.98 & \textbf{78.95} \\
Predictive   & 80.47 & \textbf{81.03} & 86.20 & \textbf{88.50} & 88.17 & \textbf{90.68} & 75.32 & \textbf{77.14} \\
Generative   & 80.86 & \textbf{82.78} & 86.81 & \textbf{89.26} & 88.74 & \textbf{91.87} & 75.87 & \textbf{78.62} \\
\midrule
Total        & 81.54 & \textbf{84.89} & 87.06 & \textbf{91.57} & 89.45 & \textbf{92.85} & 76.58 & \textbf{79.05} \\
\bottomrule
\end{tabular}
}
\end{table*}

\begin{figure}[t]
  \centering
  \includegraphics[width=\linewidth]{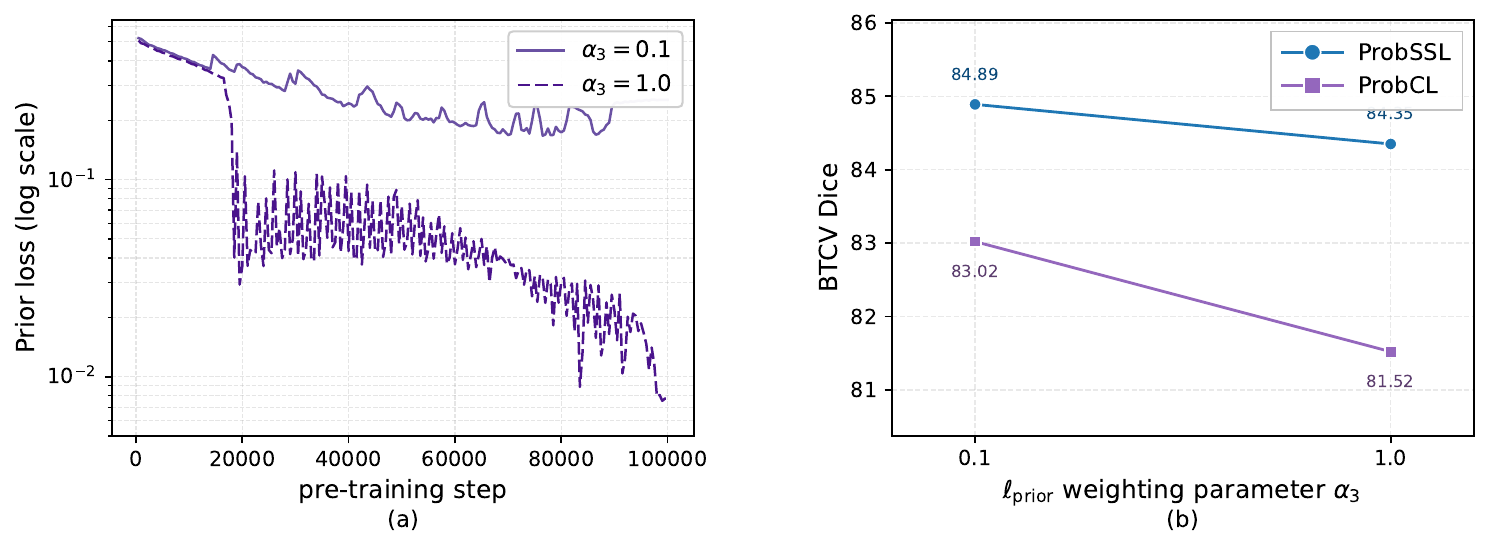}
  \caption{(a) ProbCL: logged prior term versus pre-training step for $\alpha_3=0.1$ (solid) and $\alpha_3=1.0$ (dashed). (b) BTCV mean Dice (\%) after fine-tuning versus the prior weighting coefficient $\alpha_3\in\{0.1,1.0\}$ for ProbSSL and ProbCL.}
  \label{fig:alpha3_pretrain_scalars}
\end{figure}

A further observation is that adding the prior to the sample-wise branch alone can cause a slight drop (81.63$\rightarrow$81.43). This suggests that when used in isolation, the sample-wise contrastive objective ($\mathcal{L}_{\text{SCL}}$) and the distribution-level prior regularizer can be slightly misaligned in their optimization goals. $\mathcal{L}_{\text{SCL}}$ focuses on shaping the instance-wise discriminative geometry via point-wise alignment, while the prior KL encourages the posterior to stay centered and isotropic. Without a distribution-level coupling mechanism, this prior can act as an over-regularizer that limits the flexibility of the point-wise representation.

We also examine the effect of prior regularization weight $\alpha_3$. A KL weight of $1.0$ is standard in classical ELBO settings, but in our VI-grounded ProbSSL objective the KL prior term plays the role of a regularizer whose effective strength depends on the relative scale of the self-supervised losses. We therefore tune $\alpha_3$ to avoid the prior dominating the representation geometry. We additionally retrained the model with $\alpha_3=1.0$ and fine-tuned on BTCV using the same downstream protocol as in the paper. As shown in Fig.~\ref{fig:alpha3_pretrain_scalars}(a), when $\alpha_3 = 1.0$, the (logged) prior KL quickly drops by orders of magnitude and remains very small during training, indicating that $q(\boldsymbol{z}|\boldsymbol{x})$ is being strongly pushed toward the isotropic prior $p(\boldsymbol{z})=\mathcal{N}(\boldsymbol{0},\boldsymbol{I})$. This is accompanied by degraded downstream performance in Fig.~\ref{fig:alpha3_pretrain_scalars}(b): increasing $\alpha_3$ from 0.1 to 1.0 decreases BTCV Dice for both ProbSSL (84.89$\rightarrow$84.35) and ProbCL (83.02$\rightarrow$81.52). These results suggest that $\alpha_3=1.0$ leads to strong posterior over-regularization and a collapse tendency, so we set $\alpha_3=0.1$ as a more stable choice that still provides beneficial regularization.

Table~\ref{tab:ablation_total} compares three SSL approaches in deterministic and probabilistic settings across four datasets. Probabilistic methods consistently outperform their deterministic counterparts, highlighting the benefits of probabilistic modeling. Contrastive loss achieves the highest individual probabilistic scores, indicating superior representational power. Combining all three losses results in the best overall performance, underscoring the synergy of complementary SSL strategies.

\subsubsection{Ablation Studies on Amount of Labeled Data} 

\begin{figure}[!t]
    \centering
    \includegraphics[width=\linewidth, height=0.5\linewidth]{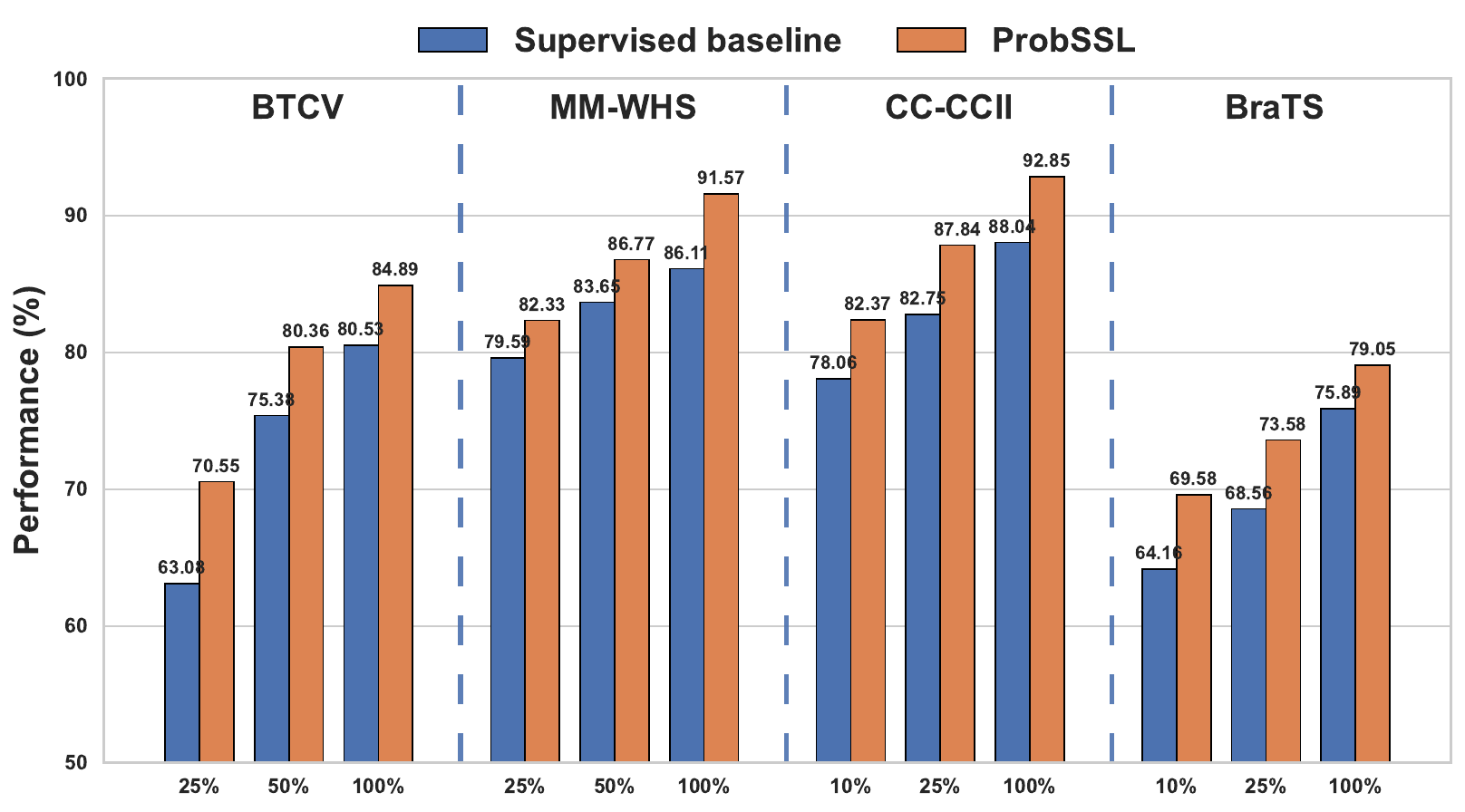}
    \caption{Performance comparison for ProbSSL and the supervised baseline at reduced label fractions (BTCV/MM-WHS: 25\%, 50\%, 100\%; CC-CCII/BraTS: 10\%, 25\%, 100\%).}
    \label{fig:few-shot}
\end{figure}

A key advantage of SSL is its reduced dependence on annotations, which is especially valuable in medical imaging, where labeling is costly and time-consuming. In our experiment, we reported downstream results on BTCV, MM-WHS, CC-CCII, and BraTS; notably, BTCV and MM-WHS already have small labeled sets (24 and 14 cases, respectively). We now report fine-tuning with only 25\% and 50\% of the labeled data on BTCV and MM-WHS, and with 10\% and 25\% on CC-CCII (2,514 training cases) and BraTS (387 training cases), comparing against our supervised baseline under the same SwinUNETR backbone.

As shown in Fig.~\ref{fig:few-shot}, the results demonstrate clear label-efficiency gains that substantiate the self-supervised claim across all four datasets. At matched label budgets, ProbSSL consistently outperforms the supervised baseline. More importantly, in several cases, ProbSSL with substantially fewer labels meets or surpasses a supervised model trained with many more labels. For example, on BTCV and MM-WHS, ProbSSL with 50\% labels matches the supervised 100\% model; on CC-CCII, ProbSSL with 25\% labels matches the supervised 100\% model. These trends show that our method achieves comparable or even better downstream accuracy with markedly fewer annotations. 

\subsubsection{Ablation Studies on Projection Head} The design of projection head, which is tied to our key novelty: it parameterizes each probabilistic embedding by predicting the mean and variance through two parallel heads, and DCL is computed on the predicted statistics. Initially, we consider a 3-layer MLP of the form \texttt{Linear-BN-ReLU} $\times 2$ + \texttt{Linear-BN} since prior SSL studies report improved performance with increasing projector depth and saturation around three layers~\cite{chen2021exploring,caron2021emerging,zbontar2021barlow}. We further replace the common \texttt{BatchNorm+ReLU} convention with \texttt{LayerNorm+GELU} to stay consistent with Swin-style transformer blocks~\cite{liu2021swin}, and avoid relying on batch statistics in our 3D pre-training regime with very small per-GPU batch sizes~\cite{ba2016layer}. While many probabilistic embedding implementations use \emph{identical} branch of heads to predict mean and variance, we adopt a slightly \emph{non-identical} design by removing the final normalization layer on the variance head, inspired by the non-identical design in~\cite{park2022probabilistic}.

We provide an ablation study where only one projection-head factor is modified at a time. We test \textbf{activation} (GELU to ReLU), \textbf{normalization} (LayerNorm to BatchNorm), and \textbf{branch design} (restoring the final normalization on the variance head so that the two heads are identical). The downstream results are summarized in Table~\ref{tab:proj_head_ablation}. Across both datasets, our projection-head design performs best.

\begin{table}[t]
\centering
\small
\caption{Ablation on projection-head design. Each variant modifies exactly one factor of our default head. Reported numbers are downstream Dice (\%).}
\label{tab:proj_head_ablation}
{\scriptsize
\begin{tabular}{lcccc}
\toprule
\textbf{Dataset} &
\textbf{ReLU (Act.)} &
\textbf{BatchNorm (Norm.)} &
\textbf{Identical heads} &
\textbf{Ours} \\
\midrule
BTCV   & 81.85 & 81.70 & 82.74 & \textbf{83.02} \\
MM-WHS & 90.31 & 89.63 & 90.45 & \textbf{90.73} \\
\bottomrule
\end{tabular}
}
\end{table}

\subsubsection{Assessing Inner Conflicts in Multi-Loss ProbSSL}
\begin{figure}[ht]
    \centering
    \includegraphics[width=\linewidth]{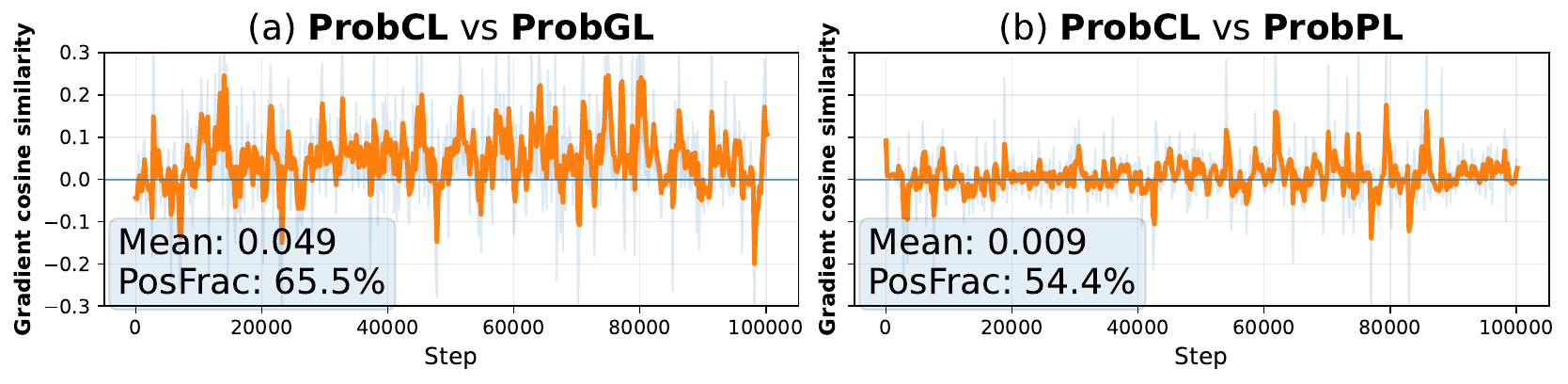}
    \caption{Gradient cosine similarity on shared parameters during pre-training between \textbf{ProbCL} and (a) \textbf{ProbGL}, (b) \textbf{ProbPL} (faint: raw; solid: EMA-smoothed). Insets report the mean and the percentage of steps with cosine similarity $>0$.}
    \label{fig:grad_cosine}
\end{figure}

    Within our multi-objective SSL framework, we emphasize that the auxiliary ProbGL/ProbPL losses are designed to be complementary to ProbCL rather than competing with it. Concretely, we diagnose inter-task interference by measuring the cosine similarity between the gradients induced by different objectives on the shared parameters (encoder and probabilistic embedding heads). As shown in Fig.~\ref{fig:grad_cosine}, the gradients between ProbCL and the auxiliary losses remain predominantly non-negative throughout training, indicating that they are largely non-conflicting and thus unlikely to cause harmful interference under the default equal-weight setting. This observation is consistent with our empirical results: adding ProbGL and ProbPL on top of ProbCL yields consistent improvements in our ablation, and the full ProbSSL achieves the best downstream performance on both BTCV and MM-WHS, suggesting no observable negative transfer in practice.

\subsubsection{Computational Cost}
We compare ProbSSL with two state-of-the-art SSL baselines, SwinUNETR~\cite{tang2022self} and VoCo~\cite{wu2024voco}, in terms of model size, computational cost (FLOPs), pretraining time, and inference time on BTCV. As summarized in Table~\ref{table: complexity_compare}, ProbSSL maintains a favorable trade-off under a shared SwinUNETR backbone. Compared with the SwinUNETR baseline, ProbSSL uses more parameters and FLOPs but introduces only modest pretraining overhead while preserving nearly identical inference time. Compared with VoCo, ProbSSL is much lighter and faster, requiring fewer parameters, FLOPs, and runtime in both pretraining and inference. Overall, these results show that ProbSSL keeps the practical efficiency of the baseline while avoiding the heavy computational burden of VoCo.

\begin{table}[!t]
\centering
\caption{Comparison of model complexity vs. performance. All models use the SwinUNETR backbone and are evaluated on the BTCV dataset.}
\label{table: complexity_compare}
\resizebox{\columnwidth}{!}{
\begin{tabular}{lcccc}
\toprule
\textbf{Method} & Params(M) & FLOPs(G) & Pretrain Time(s/epoch) & Inference Time(s/epoch)  \\
\midrule
SwinUNETR~\cite{tang2022self}  & 62.19 & 395 & 716 & 11.96 \\
VoCo~\cite{wu2024voco}         & 127.84 & 2653 & 1216  & 18.31  \\
ProbSSL                        & 87.42 & 525 & 790 & 11.97 \\
\bottomrule
\end{tabular}
}
\end{table}

\subsubsection{Failure Analysis}
\begin{figure}[!ht]
    \centering
    \includegraphics[width=\linewidth]{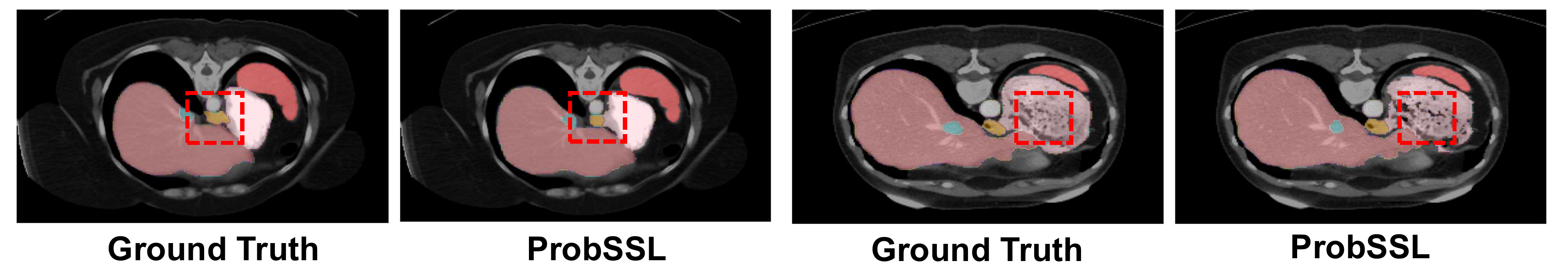}
    \caption{Qualitative discrepancy analysis. Red dashed boxes highlight localized disagreements between prediction and ground truth.}
    \label{fig:failure}
\end{figure}

We include a qualitative discrepancy panel (Fig.~\ref{fig:failure}) to make visible where our predictions differ from the reference masks. The dashed boxes highlight localized disagreements, such as small disconnected predictions and slight boundary shifts at organ interfaces. We do not speculate on causes; instead, we report these as observed inconsistencies between the prediction and the annotation. These discrepancies are spatially limited and do not alter the overall conclusions of the study, but we acknowledge them as current weaknesses and will investigate targeted remedies (e.g., removing tiny disconnected components, adding boundary-aware terms) in future work.

\subsubsection{Limitations}
Although our probabilistic objectives improve robustness in aggregate, performance can still degrade under (i) low-quality inputs (e.g., low SNR, motion or streak artifacts, inconsistent preprocessing), which inflate predictive variance and may destabilize alignment; (ii) large morphological variability and irregular shapes, where global alignment can underrepresent thin or highly nonrigid structures and increase boundary errors; (iii) long-tail and rare types, where under-represented anatomies/conditions receive weaker supervision during pretraining/fine-tuning; and (iv) cross-site/domain shifts (scanner, vendor, protocol), which are only partially accounted for by our current augmentations. In addition, (v) pretext specificity: our rotation-prediction (PL) and masked-reconstruction (GL) tasks may not be uniformly beneficial across modalities and organs, and erroneous pseudo-labels can propagate if not filtered. Finally, (vi) compute and data scale constrained the breadth of our pretraining and ablations in this manuscript.




\section{Conclusion}

In this paper, we present ProbSSL, a novel probabilistic self-supervised learning framework to learn robust representations from volumetric medical images. Our method addresses the issue of spurious invariance in medical contrastive learning by leveraging probabilistic feature representations that capture the inherent variability in medical imaging. We propose a distribution alignment loss that enables a dynamic alignment strategy where pairs with higher variability receive attenuated alignment enforcement. We provide a theoretical analysis to demonstrate that the proposed probabilistic contrastive learning loss incurs a smaller estimation error, which explicitly mitigates spurious invariance. Moreover, our framework seamlessly integrates with both predictive and generative learning methods, forming a unified probabilistic SSL approach. Extensive experiments on challenging datasets, including BTCV, MM-WHS, CC-CCII, and BraTS, demonstrate that ProbSSL consistently outperforms state-of-the-art general and medical self-supervised learning methods.

Overall, our work suggests that incorporating probabilistic modeling into the self-supervised learning framework can significantly enhance performance in medical image analysis. One limitation of our study is that computational constraints prevent us from scaling to more extensive pre-training data. Nevertheless, our results highlight the merits of this method, and we anticipate that future research will extend these findings to broader medical applications with larger datasets.

While we adopt a Gaussian probabilistic embedding for stability and closed-form distribution alignment, we acknowledge that more expressive families (e.g., mixture or heavy-tailed posteriors) may better capture extreme multimodal/outlier behaviors in certain pathology-rich settings, which we leave for future work. We adopt a diagonal covariance for efficiency, but this ignores cross-dimension correlations and may under-represent anatomical dependencies in volumetric images. This assumption can be relaxed using full covariance modeling or low-rank approximations to better capture multivariate medical structure correlations. 

For more future work, we plan to (a) report robustness under controlled corruptions and per-site/per-scanner stratifications and include uncertainty calibration metrics; (b) incorporate anatomy-aware/local contrast and boundary/topology losses to better preserve small or irregular structures; (c) adopt rare-case reweighting and curriculum/hard-example sampling to address long-tail classes; (d) evaluate lightweight domain-adaptation/test-time adaptation to reduce cross-site drift; (e) gate pseudo-labels with uncertainty thresholds and explore alternative pretexts tailored to each modality; and (f) scale pretraining to more diverse data.





\bibliographystyle{IEEEtran}
\bibliography{egbib}

\end{document}